\documentclass[10pt,journal,compsoc]{IEEEtran}

\usepackage{amsmath}
\usepackage{booktabs}
\usepackage{url}
\usepackage{graphicx}
\graphicspath{{images/}}

\newcommand{\ohta}[0]{$\text{I}_1\text{I}_2\text{I}_3$}
\newcommand{\raw}{RawFooT}
\usepackage{siunitx}

\title{Evaluating color texture descriptors under large variations of
  controlled lighting conditions}

\author{Claudio Cusano (University of Pavia)\\
\and%
Paolo Napoletano (University of Milano-Bicocca)\\
\and%
Raimondo Schettini (University of Milano-Bicocca)}

\begin{document}

\maketitle

\begin{abstract}
  The recognition of color texture under varying lighting conditions
  is still an open issue.  Several features have been proposed for
  this purpose, ranging from traditional statistical descriptors to
  features extracted with neural networks.  Still, it is not
  completely clear under what circumstances a feature performs better
  than the others.  In this paper we report an extensive comparison of
  old and new texture features, with and without a color normalization
  step, with a particular focus on how they are affected by small and
  large variation in the lighting conditions.  The evaluation is
  performed on a new texture database including 68 samples of raw food
  acquired under 46 conditions that present single and combined
  variations of light color, direction and intensity. The database
  allows to systematically investigate the robustness of texture
  descriptors across a large range of variations of imaging
  conditions.
\end{abstract}

\section{Introduction}

\begin{figure*}[tb]
\tiny
\centering
\setlength{\tabcolsep}{1pt}
  \begin{tabular}{ccccccccc}
    D65 ($\theta$=$24^{\circ}$) &
    D95 ($\theta$=$24^{\circ}$) & L27 ($\theta$=$24^{\circ}$) &
    D65 ($\theta$=$60^{\circ}$) & D95 ($\theta$=$60^{\circ}$) &
    L27 ($\theta$=$60^{\circ}$) & D65 ($\theta$=$90^{\circ}$) &
    D95 ($\theta$=$90^{\circ}$) & L27 ($\theta$=$90^{\circ}$) \\[1ex]
    \includegraphics[width=0.1\linewidth]{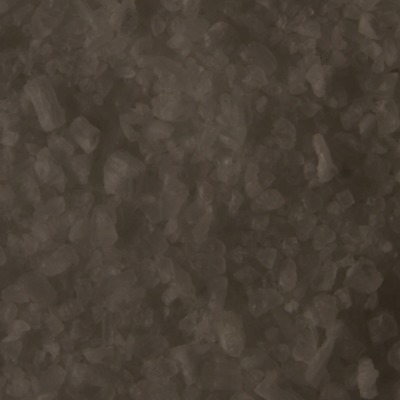} &
    \includegraphics[width=0.1\linewidth]{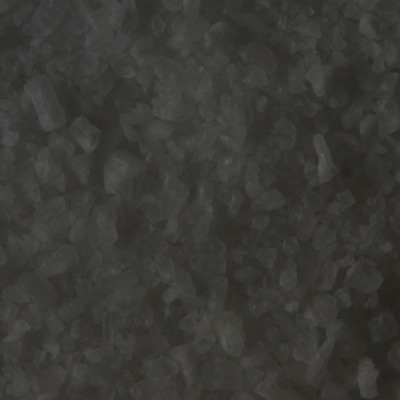} &
    \includegraphics[width=0.1\linewidth]{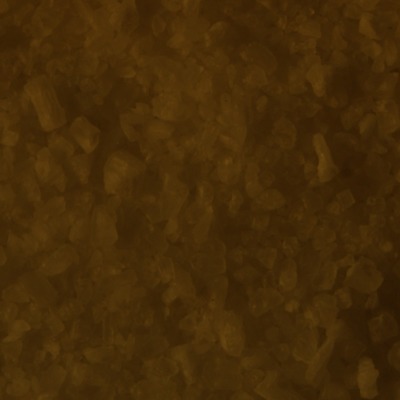} &
    \includegraphics[width=0.1\linewidth]{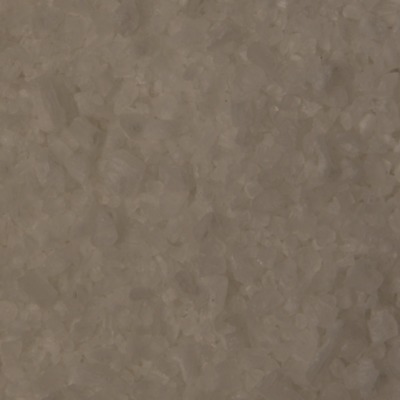} &
    \includegraphics[width=0.1\linewidth]{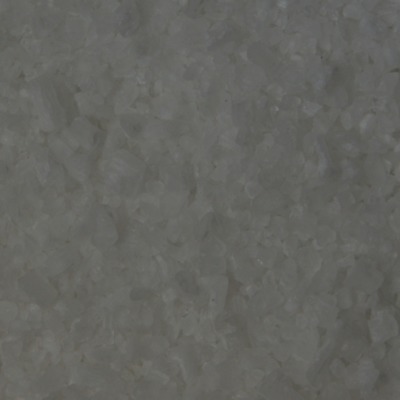} &
    \includegraphics[width=0.1\linewidth]{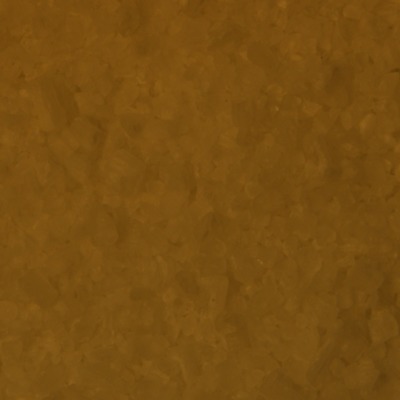} &
    \includegraphics[width=0.1\linewidth]{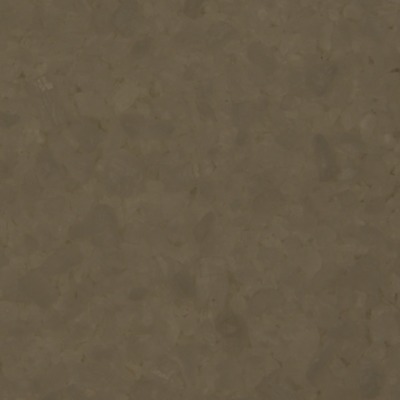} &
    \includegraphics[width=0.1\linewidth]{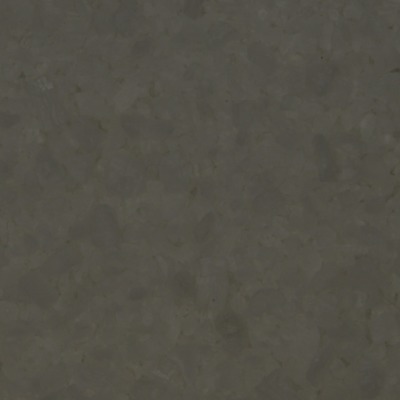} &
    \includegraphics[width=0.1\linewidth]{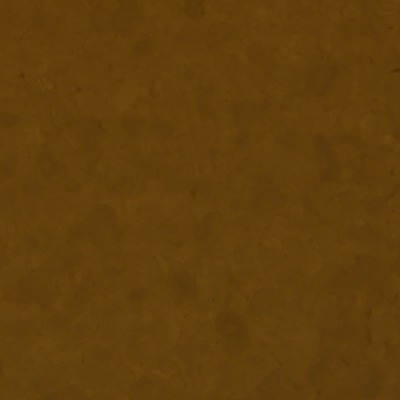} \\
    \multicolumn{9}{c}{(a)}\\[4pt]
    \includegraphics[width=0.1\linewidth]{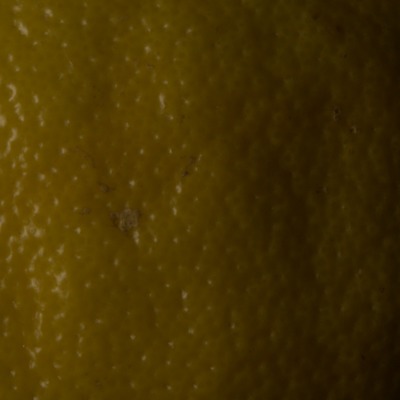} &
    \includegraphics[width=0.1\linewidth]{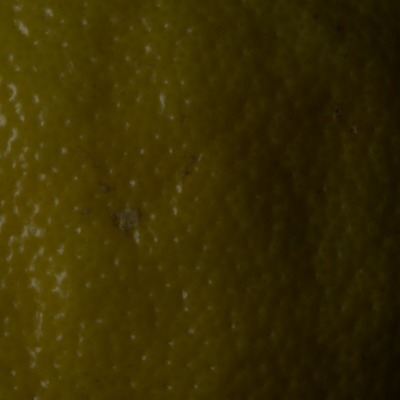} &
    \includegraphics[width=0.1\linewidth]{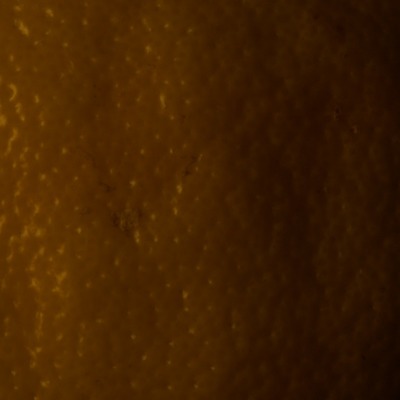} &
    \includegraphics[width=0.1\linewidth]{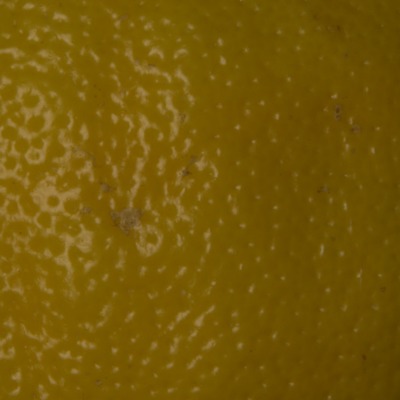} &
    \includegraphics[width=0.1\linewidth]{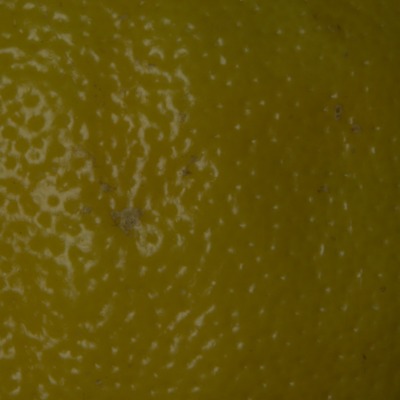} &
    \includegraphics[width=0.1\linewidth]{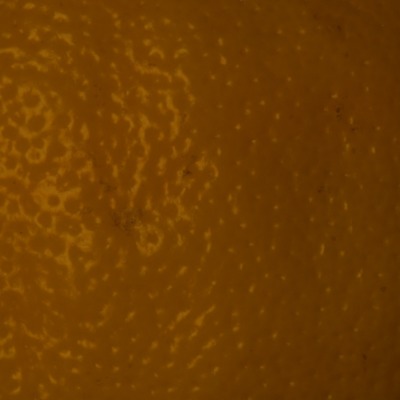} &
    \includegraphics[width=0.1\linewidth]{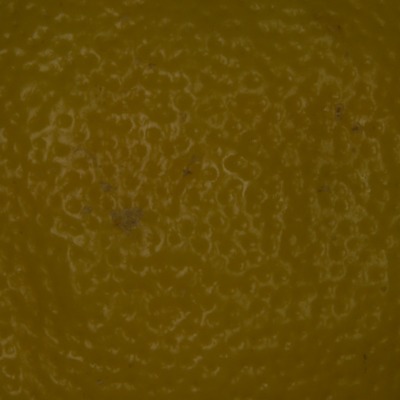} &
    \includegraphics[width=0.1\linewidth]{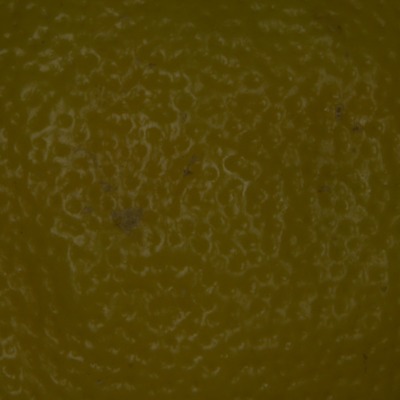} &
    \includegraphics[width=0.1\linewidth]{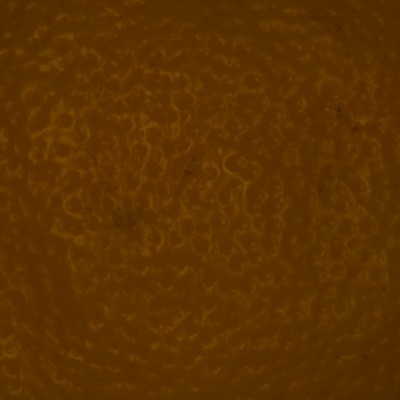} \\
    \multicolumn{9}{c}{(b)} \\
    \end{tabular}
    \caption{Examples of 2 different textures acquired under 9
      different light conditions. (a) Salt. (b) Grapefruit.}
  \label{fig:sample}
\end{figure*}

The role of color in texture classification has been widely debated in
the literature.  Despite the number and the depth of the experimental
verifications, it is still not completely clear how much and under
what circumstances color information is beneficial.  Notable examples
of this kind of analysis are the work by M\"{a}enp\"{a}\"{a} and
Pietik\"{a}inen~\cite{maenpaa2004classification}, and that by
Bianconi~\emph{et al.}~\cite{bianconi2011theoretical}.  They both
observed how color can be effective, but only in those cases where
illumination conditions do not vary too much between training and test
sets.  In fact, methods that exploit color information greatly suffer
variations in the color of the illuminant.  Under these circumstances
the best result is often achieved simply by disregarding color, that
is, by reducing all the images to gray scale.  The degree of
intra-class variability of the images in
Fig.~\ref{fig:sample} suggests why color information, if
not properly processed, can easily be deceptive.

A possible strategy to exploit color in texture classification
consists in the extraction of image features that are invariant (or at
least robust) with respect to changes in the illumination.  In scene
and object recognition the approach of specially designing invariant
features is rapidly becoming obsolete in favor of features
automatically learned from a large amount of data with methods based
on deep learning~\cite{lecun2015deep}.  It is not clear if the same is
going to happen in texture recognition as well.  A recent
work~\cite{cimpoi2015deep} suggests that a hybrid approach (local
features extracted from a convolutional neural network and then
aggregated as Fisher vectors) can be the most successful.

The availability of suitable databases of texture images is of primary
importance for the research in this field.  Therefore, in the past
several texture databases have been collected to asses the performance
of texture recognition methods under a variety of conditions.  These
databases are often focused on the exploration of the variability of
texture images under specific variations of imaging
conditions~\cite{hossain2013texture,bianconi2014appendix}, mostly
related to variations in the geometry of the acquisition setup (with
little or no variation about the characteristics of the illuminant).
For instance, several texture databases include images where the same
samples are taken from different point of views.  As a result, the
images depict the same textures taken at different scales and
orientations.  By contrast, a more recent work proposed a database of
``textures in the wild''~\cite{cimpoi2014describing} to allow texture
analysis in completely uncontrolled environments.  This approach
allows to implicitly verify the robustness against a multitude of
source of variations simultaneously.  Even though, the results on this
kind of dataset may provide a better indication about the `average'
performance of texture recognition methods in several real-world
applications, they do not allow a clear analysis of their strengths
and weaknesses for specific setups.  In fact, there are several
application domains where acquisition conditions are indeed very
controlled (medical imaging, industrial inspection\dots) and for which
the uncertainty inherent to the experimentation in the wild is a
serious liability.

In this paper we address the problem of texture classification under
controlled, large variations of lighting conditions. We have evaluated
and compared several texture and color descriptors with respect to
single and combined changes in the lighting conditions.  We selected
three classes of visual descriptors. The first class includes
\emph{traditional} (\emph{hand crafted}) descriptors specially
designed for \emph{texture analysis}. The second one includes features
that were specially designed for \emph{object recognition}.  The third
class includes those features that correspond to intermediate
representations computed by \emph{Convolutional Neural Networks}
(CNNs).

Since we addressed the problem of texture classification under varying
lighting conditions, we also investigated the use of color
normalization methods as a preprocessing, so quantifying how much
their application influences the performance of the different
descriptors.

Existing texture
databases~\cite{hossain2013texture,bianconi2014appendix} do not
include, in general, large variations of lighting conditions and, in
particular, they do not allow to evaluate the goodness of visual
descriptors with respect to both single and combined lighting
condition changes, such as only direction or temperature of the light,
direction and temperature of the light, etc. Due to these reasons, we
collected a new texture database, that we named Raw Food Texture
database (\raw). This database includes several samples of raw food
acquired under 46 conditions differing in the light color, direction
and intensity. We choose to focus on raw food because, similarly to
other natural surfaces, its color is an intrinsic property.  Therefore,
the task of classifying this kind of textures does not include the
semantic ambiguities that, instead, may arise when dealing with
artificial surfaces, where it is possible to argue that samples of
different color may be instances of the same class of ``textures''.
As far as we know the proposed database is the one featuring the
largest amount of variations in the lighting conditions, and the only
one where color, direction and intensity of light are subject to
systematic and independent changes.

In this paper we address the following issues:
\begin{itemize}
\item How effective are \emph{hand-crafted texture descriptors} when
  acquisition condition variations are so large?
\item Can \emph{object recognition descriptors} achieve high
  classification accuracy on pure texture images?
\item Do \emph{CNN-based descriptors} confirm to be powerful also on
  texture classification tasks?
\item Can \emph{CNN-based descriptors} handle large variations in
  lighting conditions?
\item Is color normalization helpful for texture analysis in case of
  changes in lighting conditions?
\end{itemize}

The rest of the paper is organized as follows: Section~\ref{sec:db}
describes the proposed \raw{} database and compares it against other
publicly available data sets; Section~\ref{sec:descriptors} reviews
the main texture descriptors in the state of art;
Section~\ref{sec:experiments} describes the experimental setup and
Section~\ref{sec:results} reports the results obtained; finally,
Section~\ref{sec:conclusions} presents our final considerations and
discusses some new directions for our future research on this topic.

\section{Texture databases}
\label{sec:db}

In the last years, different research groups developed a number of
databases of texture images ranging from natural textures to man-made
materials~\cite{hossain2013texture,bianconi2014appendix,bell2014material}. Each
database has been designed to study one or several aspects about
textures: invariance to acquisition device, invariance to lighting
conditions, invariance to image rotation or scale, 3D reconstruction,
computer graphics, classification, segmentation, etc.  The problems of
texture classification and of material recognition are closely
related.  In this paper we mainly focused on the former since the two
problems may require different
strategies~\cite{sharan2013recognizing}.

We considered the most important texture databases that have been
presented in the
literature~\cite{hossain2013texture,bianconi2014appendix}, and we
compiled the Table~\ref{tab:db_comparison} where we have highlighted
the most important features of each database. Amongst the features,
the most important are those related to the acquisition setup, such
as: \emph{illumination conditions}, \emph{sensor's angle}, \emph{image
  rotation}, \emph{scaling} and \emph{color of the illuminant}. In
particular, we highlighted four different sources of variations in
illumination conditions. One related to the \emph{direction} of the
light, one to the \emph{intensity} of the light, another to the
\emph{color temperature} of the light, and the fourth related to a
mixture of variations, such as: temperature and direction, temperature
and intensity, etc.  As you can see from the table, several databases
consider a mixture of variations. The most notable is the OuTex
database~\cite{outex} which is, in fact, the most used in the study of
descriptors invariant to lighting
conditions~\cite{maenpaa2004classification}.  The OuTex collection
includes the OuTex-14 test suite that contains images that depict
textures of 68 different classes acquired under three different light
sources, each positioned differently: the \SI{2856}{K} incandescent
CIE A, the \SI{2300}{K} horizon sunlight and the \SI{4000}{K}
fluorescent TL84.

\begin{table*}[tb]
  \caption{Main features of existing texture
    databases~\cite{dbtexture,dbtextureappendix}.  The filled circle
    indicates that the given feature is present, the empty circle
    indicates its absence, the minus sign indicates that information
    on that feature is not available.}
\centering
\label{tab:tools}
\tiny
\def\arraystretch{1.6}%
\begin{tabular}{p{1.0cm}p{0.7cm}p{0.95cm}p{0.8cm}p{0.8cm}p{0.82cm}%
  p{0.82cm}p{0.82cm}p{0.82cm}p{0.88cm}p{1.1cm}p{0.8cm}p{1.2cm}}
  \toprule
  Database &
  Type of surface & 
  \#Classes \#Images &
  Image size &
  Color repr. &
  Rotation & 
  Scaling &
  Light directions &
  Light intensity &
  Light temperature &
  Lighting variability &
  Sensor's angle &
  Camera sensor \\
\midrule
Brodatz ~\cite{brodatz1966textures} & natural, artificial &
112 classes 112 images & 512$\times$512 & gray-scale &
$\bullet$ & $\circ$ & $\circ$ &$\circ$&$\circ$ &$\bullet$ controlled unknown&scanned&$-$\\
CUReT~\cite{Curet} 	& natural, artificial & 61 classes over \num{14000} images & 512$\times$512 & color RGB & $\bullet$ &$\circ$ & $\bullet$ 55& $\circ$ &$\circ$&$\circ$&7 positions&3 separate CCD\\
VisTex~\cite{vistex}	&natural, artificial  & 54 classes 864 images & 512$\times$512  & $-$  &$\circ$ & $\circ$  & $\bullet$   unknown directions  & $\circ$  & $\circ$  & $\bullet$ uncontrolled:
daylight,
fluorescent,
incandescent&frontal, oblique&$-$\\
MeasTex~\cite{meastex}	& natural, artificial & 4 classes  944 images & 512$\times$512  & gray-scale& $\bullet$ &$\circ$ &$\circ$ & $\circ$ & $\circ$ &  $\bullet$ daylight direct and indirect,
flash&frontal, oblique&scanned from 35mm film\\
PhoTex~\cite{photex}			& rough surfaces  & 64 classes & 1280$\times$1024 & gray-scale & $\bullet$ & $\circ$ & $\bullet$ 4 & $\circ$ & $\circ$ &$\circ$&12 positions&Vosskuhler CCD 1300LN\\
OuTex~\cite{outex}	& natural, artificial &320 classes \num{51840} images&516$\times$716  & color RGB & $\bullet$ & $\bullet$ & $\circ$ & $\circ$ & $\circ$ &
$\bullet$ fluorescent, tl84,
incandescent at 3 positions&frontal&3 CCD, Sony DXC-755P\\
DynTex~\cite{dyntex} 						&natural, artificial  & 36 classes 345 sequences & 720$\times$576 25fps & color RGB & $\bullet$ & $\circ$ & $\circ$ & $\circ$ & $\circ$ & $\circ$ &$-$&3 CCD, Sony DCR-TRV890E TRV900E\\
UIUC	~\cite{uiuc}					& natural, artificial &25 classes 1000 images  & 640$\times$480 & gray-scale& $\circ$ & $\bullet$ & $\circ$ & $\circ$ & $\circ$ &$\circ$&various&$-$\\
KTH-TIPS2~\cite{kth-tips2} 							& natural, food, artificial &44 classes 4608 images & 1280$\times$960 & color RGB & $\bullet$ & $\bullet$ & $\bullet$ 3& $\circ$ & $\circ$ &$\bullet$ incandescent, fluorescent&frontal, oblique&Olympus C-3030ZOOM\\
ALOT	~\cite{alot}							& natural, food, artificial & 250 classes, over \num{27500} & &color RGB & $\bullet$ & $\circ$ & $\bullet$ 5& $\circ$ & $\bullet$ 2 & $\circ$ &4 positions&Foveon X3 3CMOS\\
Mondial Marmi~\cite{mondialmarmi}				& granite  & 12 classes  & 544$\times$544 & color RGB & $\bullet$ & $\circ$ & $\circ$ & $\circ$ & $\circ$ &$\bullet$ white HB LEDs&frontal&Samsung S850\\
STex~\cite{stex}					& natural, artificial & 476 classes & 1024$\times$1024  & color RGB & $\circ$ & $\circ$ & $\circ$ & $\circ$ &$\circ$ &$\circ$&$\circ$&$-$\\
USPTex~\cite{usptex} & natural, artificial & 191 classes, 2292 images &  512$\times$384 & color RGB & $\circ$ & $\circ$ & $\circ$ & $\circ$ & $\circ$& $\circ$ &$-$&$-$\\
Proposed \raw 		& natural, food & 68 classes, 3128 images  & 800$\times$800 & color RGB & $\circ$ & $\circ$ & $\bullet$ 9 & $\bullet$ 5& $\bullet$ 18 &$\bullet$ 12&frontal&Canon 40D\\[1ex]
\bottomrule
\end{tabular}
\label{tab:db_comparison}
\end{table*}

Few databases consider variations of light direction, intensity or
temperature separately. In particular, the only database that provides
a good number of this kind of variations is the ALOT
database~\cite{alot}. This collection provides 250 classes of textures
acquired under several conditions obtained combining five illumination
directions (at \SI{3075}{K}) and one semi-hemispherical illumination
(at \SI{2175}{K}). Each object was recorded with only one out of five
lights turned on, yielding five different illumination angles. One
image is recorded with all lights turned on, yielding a sort of
hemispherical illumination. All the images are acquired by four
cameras positioned differently.

As far as we know, no publicly available texture database has been
designed to assess the performance in texture classification under a
broad range of variations in the illumination color, direction and
intensity.  This is why we collected the Raw Food Texture database.

\subsection{The Raw Food Texture database (\raw)}
The Raw Food Texture database (\raw) has been specially designed to
investigate the robustness of descriptors and classification methods
with respect to variations in the lighting conditions, with a
particular focus on variations in the color of the illuminant.  The
database includes images of samples of textures, acquired under 46
lighting conditions which may differ in the light direction, in the
illuminant color, in its intensity, or in a combination of these
factors.

\begin{figure*}[tb]
\tiny
\centering
\setlength{\tabcolsep}{1.5pt}
  \begin{tabular}{*{12}{l}}
     \includegraphics[width=0.075\linewidth]{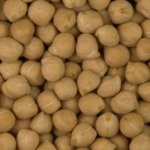} &
     \includegraphics[width=0.075\linewidth]{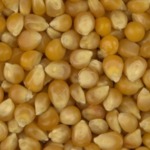} &
     \includegraphics[width=0.075\linewidth]{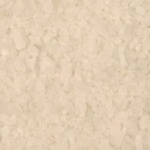} &
     \includegraphics[width=0.075\linewidth]{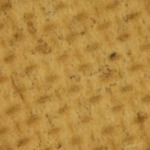} &
     \includegraphics[width=0.075\linewidth]{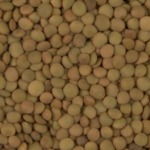} &
     \includegraphics[width=0.075\linewidth]{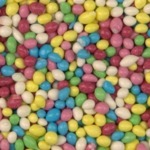} &
     \includegraphics[width=0.075\linewidth]{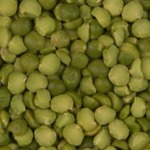} &
     \includegraphics[width=0.075\linewidth]{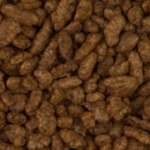} &
     \includegraphics[width=0.075\linewidth]{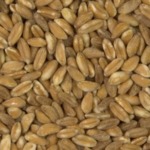} &
     \includegraphics[width=0.075\linewidth]{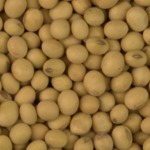} &
     \includegraphics[width=0.075\linewidth]{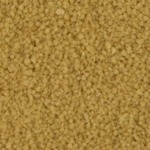} &
     \includegraphics[width=0.075\linewidth]{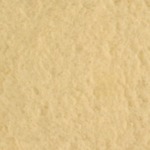} \\
     (1) chickpeas & (2) corn & (3) salt & (4) cookie &
     (5) lentils & (6) candies & (7) green peas & (8) puffed rice &
     (9) spelt & (10) white peas & (11) cous cous & (12) sliced bread \\
     \includegraphics[width=0.075\linewidth]{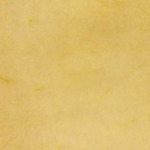} &
     \includegraphics[width=0.075\linewidth]{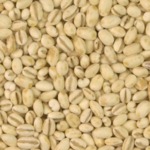} &
     \includegraphics[width=0.075\linewidth]{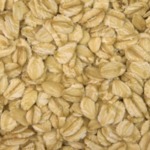} &
     \includegraphics[width=0.075\linewidth]{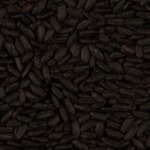} &
     \includegraphics[width=0.075\linewidth]{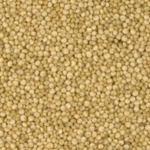} &
     \includegraphics[width=0.075\linewidth]{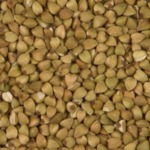} &
     \includegraphics[width=0.075\linewidth]{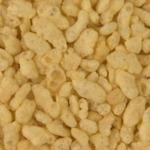} &
     \includegraphics[width=0.075\linewidth]{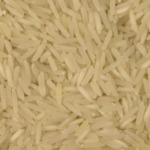} &
     \includegraphics[width=0.075\linewidth]{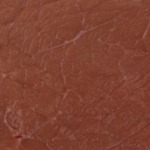} &
     \includegraphics[width=0.075\linewidth]{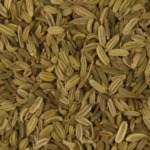} &
     \includegraphics[width=0.075\linewidth]{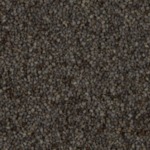} &
     \includegraphics[width=0.075\linewidth]{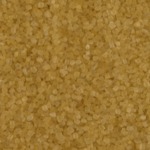} \\
     (13) apple slice & (14) pearl barley &
     (15) oat & (16) black rice & (17) quinoa & (18) buckwheat &
     (19) puffed rice & (20) basmati rice & (21) steak &
     (22) fennel seeds & (23) poppy seeds & (24) brown sugar \\
     \includegraphics[width=0.075\linewidth]{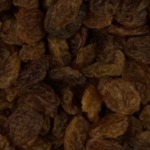} &
     \includegraphics[width=0.075\linewidth]{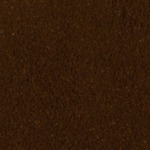} &
     \includegraphics[width=0.075\linewidth]{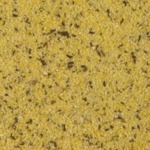} &
     \includegraphics[width=0.075\linewidth]{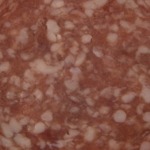} &
     \includegraphics[width=0.075\linewidth]{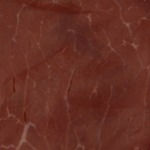} &
     \includegraphics[width=0.075\linewidth]{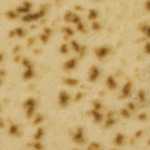} &
     \includegraphics[width=0.075\linewidth]{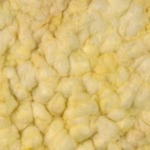} &
     \includegraphics[width=0.075\linewidth]{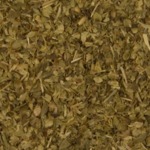} &
     \includegraphics[width=0.075\linewidth]{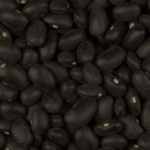} &
     \includegraphics[width=0.075\linewidth]{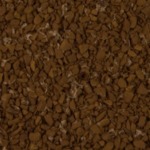} &
     \includegraphics[width=0.075\linewidth]{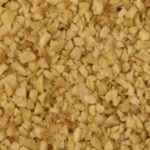} &
     \includegraphics[width=0.075\linewidth]{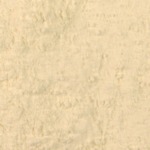} \\
     (25) sultana & \makebox[0pt][l]{(26) coffee powder} &
     (27) polenta flour &
     (28) salami & \makebox[0pt][l]{(29) air-cured beef} &
     (30) flatbread & (31) corn crackers &
     (32) oregano & (33) black beans &
     \makebox[0pt][l]{(34) soluble coffee} &
     \makebox[0pt][l]{(35) hazelnut grain} & (36) flour \\
     \includegraphics[width=0.075\linewidth]{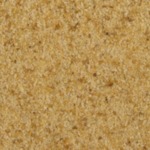} &
     \includegraphics[width=0.075\linewidth]{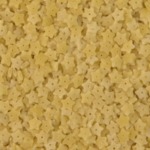} &
     \includegraphics[width=0.075\linewidth]{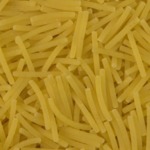} &
     \includegraphics[width=0.075\linewidth]{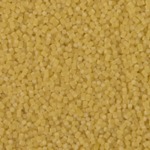} &
     \includegraphics[width=0.075\linewidth]{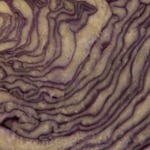} &
     \includegraphics[width=0.075\linewidth]{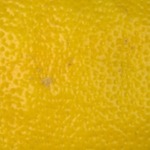} &
     \includegraphics[width=0.075\linewidth]{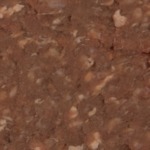} &
     \includegraphics[width=0.075\linewidth]{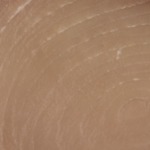} &
     \includegraphics[width=0.075\linewidth]{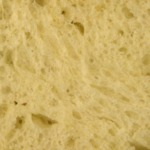} &
     \includegraphics[width=0.075\linewidth]{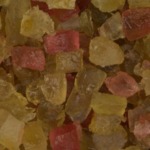} &
     \includegraphics[width=0.075\linewidth]{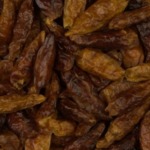} &
     \includegraphics[width=0.075\linewidth]{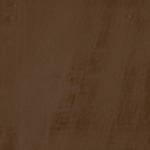} \\
     (37) bread crumbs &
     (38) pasta (stars) & (39) cut spaghetti & (40) pastina &
     (41) red cabbage & (42) grapefruit &
     (43) hamburger & (44) swordfish & (45) bread &
     (46) candied fruit & (47) chili pepper &
     \makebox[0pt][l]{(48) milk chocolate} \\
     \includegraphics[width=0.075\linewidth]{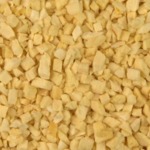} &
     \includegraphics[width=0.075\linewidth]{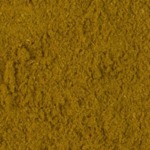} &
     \includegraphics[width=0.075\linewidth]{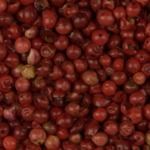} &
     \includegraphics[width=0.075\linewidth]{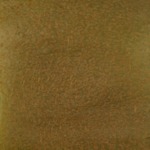} &
     \includegraphics[width=0.075\linewidth]{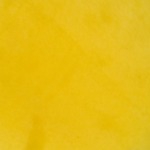} &
     \includegraphics[width=0.075\linewidth]{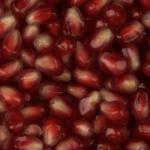} &
     \includegraphics[width=0.075\linewidth]{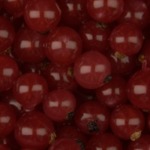} &
     \includegraphics[width=0.075\linewidth]{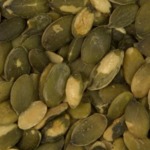} &
     \includegraphics[width=0.075\linewidth]{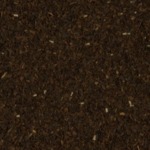} &
     \includegraphics[width=0.075\linewidth]{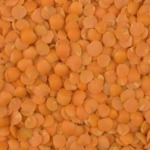} &
     \includegraphics[width=0.075\linewidth]{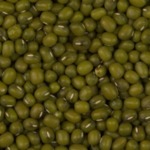} &
     \includegraphics[width=0.075\linewidth]{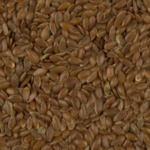} \\
     (49) garlic grain & (50) curry & (51) pink pepper &
     (52) kiwi & (53) mango & (54) pomegranate & (55) currant &
     (56) \makebox[0pt][l]{pumpkin seeds} &
     (57) tea & (58) red lentils & (59) green adzuki &
     (60) linseeds \\
     \includegraphics[width=0.075\linewidth]{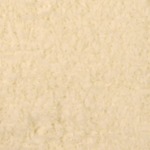} &
     \includegraphics[width=0.075\linewidth]{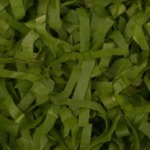} &
     \includegraphics[width=0.075\linewidth]{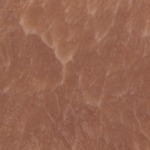} &
     \includegraphics[width=0.075\linewidth]{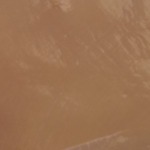} &
     \includegraphics[width=0.075\linewidth]{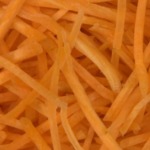} &
     \includegraphics[width=0.075\linewidth]{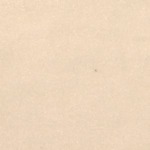} &
     \includegraphics[width=0.075\linewidth]{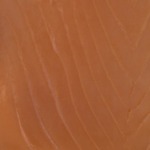} &
     \includegraphics[width=0.075\linewidth]{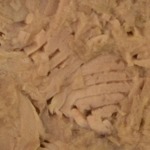} \\
     \makebox[0pt][l]{(61) coconut flakes} & (62) chicory &
     (63) pork loin & \makebox[0pt][l]{(64) chicken breast} &
     (65) carrots &
     (66) sugar & (67) salmon & (68) tuna
   \end{tabular}
   \caption{Overview of the 68 classes included in the Raw Food
     Texture database. For each class it is shown the image taken
     under D65 at direction $\theta$=$24^{\circ}$.}
  \label{fig:db_overview}
\end{figure*}

Psycho-physical studies~\cite{poirson1996pattern-color} suggest that,
in the human visual system, color and pattern information are
processed separately.  However, it has been observed that their
combination can be very effective for texture classification.  For
certain classes of materials the two kind of information are clearly
independent (e.g.~fabrics and other artificial materials).  For this
reason, we considered samples of texture where the relationship
between pattern information and color has not been explicitly
designed.  Our classes correspond to 68 samples of raw food, including
various kind of meat, fish, cereals, fruit etc.  Therefore, the whole
database includes $68 \times 46 = 3128$ images.
Fig.~\ref{fig:db_overview} shows an image of each sample.

\begin{figure}[tb]
  \begin{center}
    \includegraphics[width=0.9\linewidth]{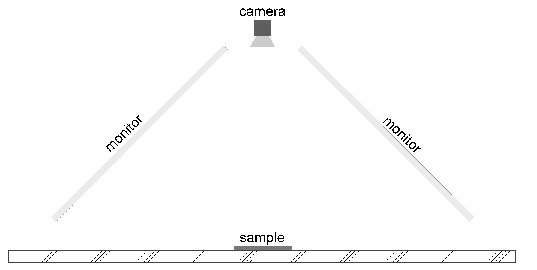}
  \end{center}
  \caption{The setup used to acquire the Raw Food Texture database.}
  \label{fig:acquisition_setup}
\end{figure}

Pictures have been acquired in a dark room with a Canon EOS 40D DSLR
camera.  The camera was placed 48cm above the sample to be acquired,
with the optical axis perpendicular to the surface of the sample.  The
lenses used had a focal length of 85mm, and a camera aperture of
f/11.3; each picture has been taken with four seconds of exposition
time.  As illuminants, we used a pair of monitors (22 inches Samsung
SyncMaster LED monitor) positioned above the sample and tilted by 45
degrees, with about 20cm of space between their upper edges to make
room for the camera.  By illuminating different regions of the
monitors, and by using different colors (inspired
by~\cite{khan2013towards}) we simulated natural and artificial
illuminants coming from different directions and at various intensity
levels.  The two monitors have been colorimetrically characterized
using a X-Rite i1 spectral colorimeter, in such a way that the device
RGB coordinates can be used to accurately render the desired
chromaticites.  Daylight at \SI{6500}{K} (D65) has been specified as a
white point.  Fig.~\ref{fig:acquisition_setup} shows the setup used
for the acquisitions.

For each sample a program of 46 shots has been followed:\\
\textbf{intensity variations}: four shots have been taken while
illuminating the whole monitors with neutral light (D65) at different
levels of intensity (100\%, 75\%, 50\% and 25\% of the maximum
achievable level);\\
\textbf{light direction:} nine shots have been taken with the light
(D65) coming from different angles.  In the first eight of these shots
only a band covering 40\% of a single monitor has been lit.  The angle
between the direction of the light coming from the center of the
illuminated band and the surface of the sample are 24, 30, 36, 42, 48,
54, 60, and 66 degrees.  For the last shot two bands covering the
upper 20\% of each monitor have been lit (since on average the light
comes exactly from above the sample, we count it as an angle of 90
degrees). \\
\textbf{Daylight:} 12 shots have been taken while simulating natural
daylight at different color temperatures. To do so, given a color
temperature $T$ we applied the following equations to obtain the
corresponding $xy$ chromaticities:
\begin{equation}
  \begin{split}
    x &= a_0 + a_1 \frac{10^3}{T} + a_2 \frac{10^6}{T^2} +
    a_3 \frac{10^9}{T^3}, \\
    y &= -3x^2 + 2.87 x - 0.275,
  \end{split}
\end{equation}
where $a_0 = 0.244063$, $a_1 = 0.09911$, $a_2 = 2.9678$,
$a_3 = -4.6070$ if \SI{4000}{K} $\leq T \leq$ \SI{7000}{K}, and
$a_0 = 0.23704$, $a_1 = 0.24748$, $a_2 = 1.9018$, $a_3 = -2.0064$ if
\SI{7000}{K} $< T \leq$ \SI{25000}{K}~\cite{wyszecki1982color}.
Chromaticities have been converted in the RGB space with a scaling of
the color channels ensuring that their largest value is 255.  We
considered 12 color temperatures in the range from \SI{4000}{K} to
\SI{9500}{K} with a step of \SI{500}{K} (we will refer to these as
D40, D45, \dots, D95).  The whole
monitors have been lit during these shots. \\
\textbf{Indoor illumination:} six shots have been taken while
simulating an artificial light with a color temperature of
\SI{2700}{K}, \SI{3000}{K}, \SI{4000}{K}, \SI{5000}{K}, \SI{5700}{K}
and \SI{6500}{K} on the two whole monitors.  We considered LED lights
produced by OSRAM and we computed the corresponding RGB values
starting from the chromaticities indicated in the data sheets from the
producer's web site\footnote{\url{http://www.osram-os.com}}.  We will
refer to these as L27, L30, \dots, L65. \\
\textbf{Color and direction:} nine shots have been taken by varying
both the color and the direction of the illuminant.  The combinations
of three colors (D65, D95 and L27) and of three
directions (24, 60 and 90 degrees) have been considered. \\
\textbf{Multiple illuminants:} three shots have been taken while the
sample is illuminated by two illuminants with different colors (D65,
D95 or L27).  Bands covering the lower 40\% of both the monitors
have been lit, using two different colors on the two monitors. \\
\textbf{Primary colors:} three shots have been taken under pure red,
green and blue illuminants.

\begin{figure*}[tb]
\tiny
\centering
\setlength{\tabcolsep}{1pt}
  \begin{tabular}{cccccccccccc}
    \includegraphics[width=0.075\linewidth]{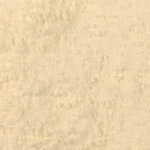} &
    \includegraphics[width=0.075\linewidth]{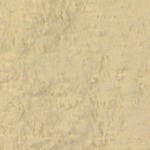} &
    \includegraphics[width=0.075\linewidth]{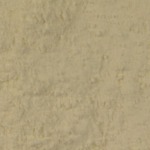} &
    \includegraphics[width=0.075\linewidth]{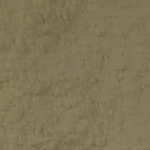} &
    \includegraphics[width=0.075\linewidth]{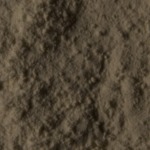} &
    \includegraphics[width=0.075\linewidth]{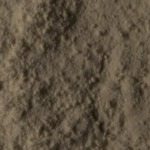} &
    \includegraphics[width=0.075\linewidth]{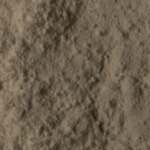} &
    \includegraphics[width=0.075\linewidth]{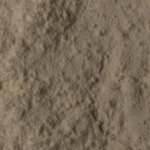} &
    \includegraphics[width=0.075\linewidth]{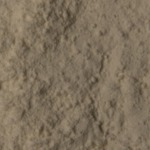} &
    \includegraphics[width=0.075\linewidth]{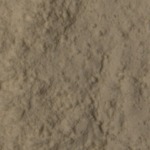} &
    \includegraphics[width=0.075\linewidth]{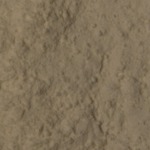} &
    \includegraphics[width=0.075\linewidth]{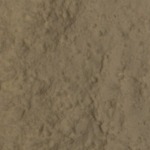} \\
    \includegraphics[width=0.075\linewidth]{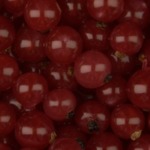} &
    \includegraphics[width=0.075\linewidth]{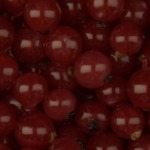} &
    \includegraphics[width=0.075\linewidth]{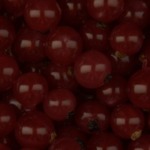} &
    \includegraphics[width=0.075\linewidth]{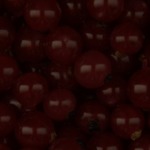} &
    \includegraphics[width=0.075\linewidth]{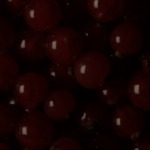} &
    \includegraphics[width=0.075\linewidth]{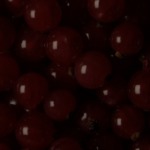} &
    \includegraphics[width=0.075\linewidth]{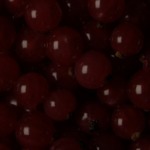} &
    \includegraphics[width=0.075\linewidth]{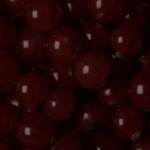} &
    \includegraphics[width=0.075\linewidth]{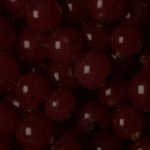} &
    \includegraphics[width=0.075\linewidth]{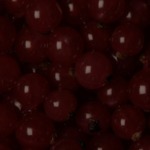} &
    \includegraphics[width=0.075\linewidth]{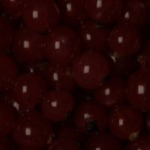} &
    \includegraphics[width=0.075\linewidth]{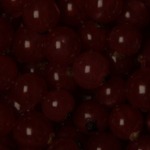}\\
    D65 (I=100\%)&D65 (I=75\%)&D65 (I=50\%)&D65 (I=25\%) &
    D65 ($\theta$=$24^{\circ}$)&D65 ($\theta$=$30^{\circ}$) &
    D65 ($\theta$=$36^{\circ}$)&D65 ($\theta$=$42^{\circ}$) &
    D65 ($\theta$=$48^{\circ}$)&D65 ($\theta$=$54^{\circ}$) &
    D65 ($\theta$=$60^{\circ}$)& D65 ($\theta$=$66^{\circ}$) \\[2ex]
    \includegraphics[width=0.075\linewidth]{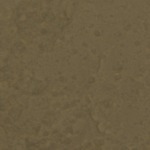} &
    \includegraphics[width=0.075\linewidth]{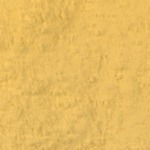} &
    \includegraphics[width=0.075\linewidth]{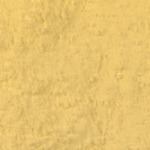} &
    \includegraphics[width=0.075\linewidth]{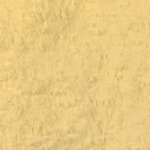} &
    \includegraphics[width=0.075\linewidth]{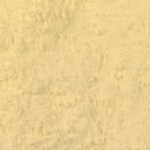} &
    \includegraphics[width=0.075\linewidth]{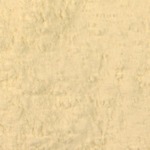} &
    \includegraphics[width=0.075\linewidth]{0036-19} &
    \includegraphics[width=0.075\linewidth]{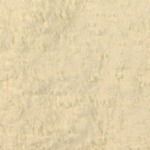} &
    \includegraphics[width=0.075\linewidth]{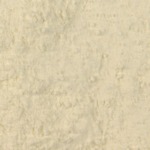} &
    \includegraphics[width=0.075\linewidth]{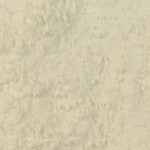} &
    \includegraphics[width=0.075\linewidth]{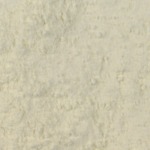} &
    \includegraphics[width=0.075\linewidth]{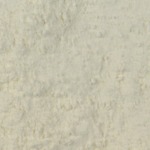} \\
    \includegraphics[width=0.075\linewidth]{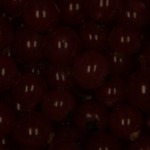} &
    \includegraphics[width=0.075\linewidth]{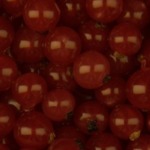} &
    \includegraphics[width=0.075\linewidth]{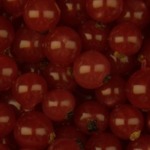} &
    \includegraphics[width=0.075\linewidth]{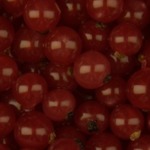} &
    \includegraphics[width=0.075\linewidth]{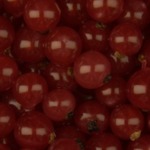} &
    \includegraphics[width=0.075\linewidth]{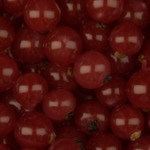} &
    \includegraphics[width=0.075\linewidth]{0055-19} &
    \includegraphics[width=0.075\linewidth]{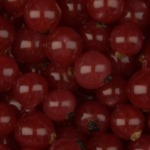} &
    \includegraphics[width=0.075\linewidth]{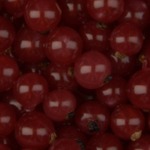} &
    \includegraphics[width=0.075\linewidth]{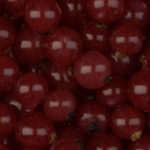} &
    \includegraphics[width=0.075\linewidth]{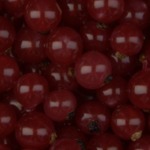} &
    \includegraphics[width=0.075\linewidth]{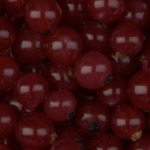} \\
    D65 ($\theta$=$90^{\circ}$)&D40&D45&D50&D55&D60&D65&
    D70&D75&D80&D85&D90\\[2ex]
    \includegraphics[width=0.075\linewidth]{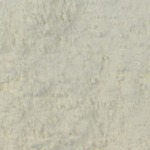} &
    \includegraphics[width=0.075\linewidth]{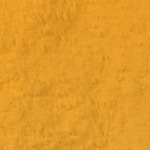} &
    \includegraphics[width=0.075\linewidth]{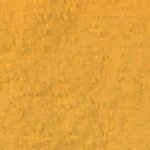} &
    \includegraphics[width=0.075\linewidth]{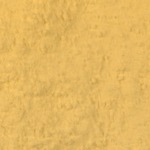} &
    \includegraphics[width=0.075\linewidth]{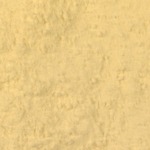} &
    \includegraphics[width=0.075\linewidth]{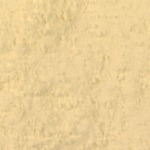} &
    \includegraphics[width=0.075\linewidth]{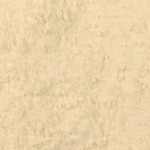} &
    \includegraphics[width=0.075\linewidth]{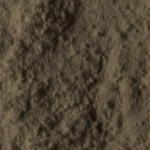} &
    \includegraphics[width=0.075\linewidth]{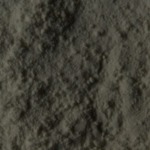} &
    \includegraphics[width=0.075\linewidth]{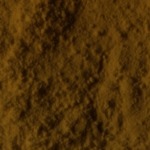} &
    \includegraphics[width=0.075\linewidth]{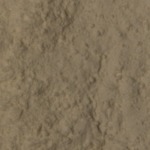} &
    \includegraphics[width=0.075\linewidth]{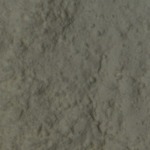} \\
    \includegraphics[width=0.075\linewidth]{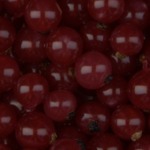} &
    \includegraphics[width=0.075\linewidth]{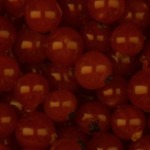} &
    \includegraphics[width=0.075\linewidth]{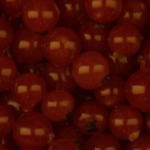} &
    \includegraphics[width=0.075\linewidth]{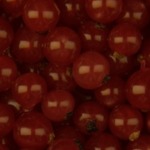} &
    \includegraphics[width=0.075\linewidth]{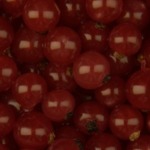} &
    \includegraphics[width=0.075\linewidth]{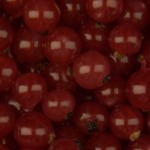} &
    \includegraphics[width=0.075\linewidth]{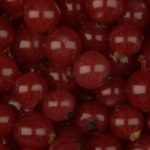} &
    \includegraphics[width=0.075\linewidth]{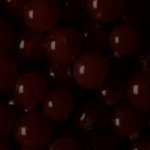} &
    \includegraphics[width=0.075\linewidth]{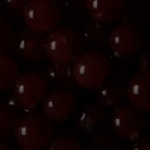} &
    \includegraphics[width=0.075\linewidth]{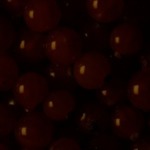} &
    \includegraphics[width=0.075\linewidth]{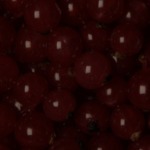} &
    \includegraphics[width=0.075\linewidth]{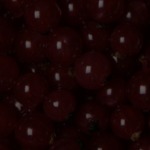} \\
    D95&L27&L30&L40
    &L50&L57&L65&D65 ($\theta$=$24^{\circ}$)&
    D95 ($\theta$=$24^{\circ}$)&L27 ($\theta$=$24^{\circ}$) &
    D65 ($\theta$=$60^{\circ}$)&D95 ($\theta$=$60^{\circ}$) \\ [2ex]
    \includegraphics[width=0.075\linewidth]{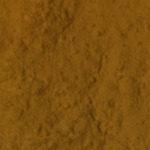} &
    \includegraphics[width=0.075\linewidth]{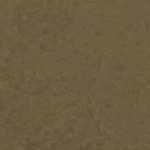} &
    \includegraphics[width=0.075\linewidth]{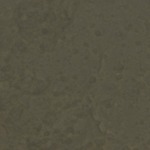} &
    \includegraphics[width=0.075\linewidth]{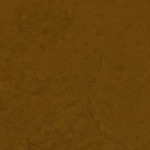} &
    \includegraphics[width=0.075\linewidth]{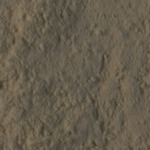} &
    \includegraphics[width=0.075\linewidth]{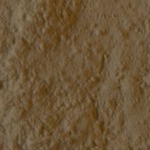} &
    \includegraphics[width=0.075\linewidth]{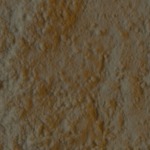} &
    \includegraphics[width=0.075\linewidth]{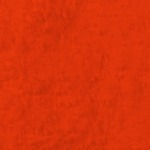} &
    \includegraphics[width=0.075\linewidth]{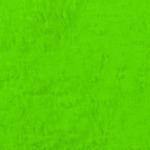} &
    \includegraphics[width=0.075\linewidth]{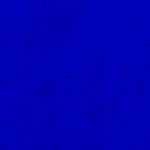} &&\\
    \includegraphics[width=0.075\linewidth]{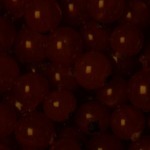} &
    \includegraphics[width=0.075\linewidth]{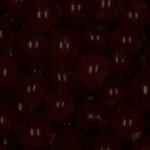} &
    \includegraphics[width=0.075\linewidth]{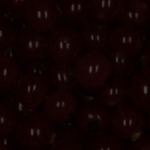} &
    \includegraphics[width=0.075\linewidth]{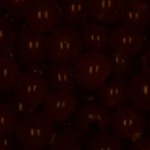} &
    \includegraphics[width=0.075\linewidth]{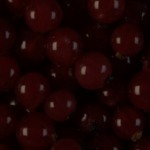} &
    \includegraphics[width=0.075\linewidth]{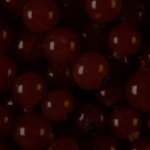} &
    \includegraphics[width=0.075\linewidth]{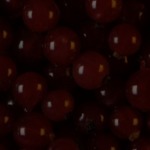} &
    \includegraphics[width=0.075\linewidth]{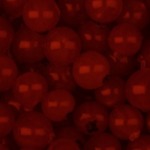} &
    \includegraphics[width=0.075\linewidth]{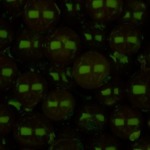} &
    \includegraphics[width=0.075\linewidth]{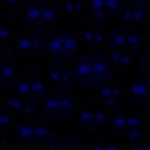} &&\\
    L27 ($\theta$=$60^{\circ}$)&D65 ($\theta$=$90^{\circ}$)&
    D95 ($\theta$=$90^{\circ}$)&L27 ($\theta$=$90^{\circ}$)&
    D65-D95&D65-L27&D95-L27&Red&Green&Blue&&\\ [2ex]
\end{tabular}
\caption{Overview of the 46 lighting conditions in the Raw Food
  Texture database: the top rows represent the \emph{flour} class
  while bottom rows represent the \emph{currant} class. }
  \label{fig:db_setup_overview}
\end{figure*}

Each \num{3944 x 2622} picture in the camera space has been converted
to standard sRGB and the final texture images have been obtained by
cropping the central region of \num{800 x 800} pixels.
Fig.~\ref{fig:db_setup_overview} shows the 46 shots taken for two of
the 68 samples. To allow the estimate of the illuminants we have
carried out the program of 46 shots of a 24 squares Macbeth
ColorChecker.

\section{Texture Descriptors}
\label{sec:descriptors}

A huge variety of texture descriptors have been proposed in the
literature.  These were traditionally divided into statistical,
spectral, structural and hybrid approach~\cite{mirmehdi2009handbook}.
Among traditional methods the most popular are probably those based on
histograms, Gabor filters~\cite{Bianconi2007}, cooccurrence
matrices~\cite{haralick1979statistical}, and Local Binary
Patterns~\cite{ojala2002multiresolution}.  These descriptors display
different strengths and weaknesses in particular concerning their
invariance with respect to the acquisition conditions.

Traditional descriptors are often designed to capture texture
information in uncluttered images taken under controlled conditions.
To address those cases where the conditions cannot be controlled, a
few attempts have been made to adapt features used for scene or object
recognition to the domain of texture classification.  For instance,
Sharan et al.~\cite{sharan2013recognizing} used SIFT and HOG
descriptors for material classification, while Sharma et
al.~\cite{sharma2012local} used a variation of the Fisher Vector
approach for texture and face classification.  Cimpoi et
al.~\cite{cimpoi2014describing} shown how SIFT descriptors aggregated
with the improved Fisher vector method greatly outperform previous
descriptors in the state of the art on a variety of texture
classification tasks, including the classification of ``textures in
the wild.''

Following the trend in image recognition, features extracted from
Convolutional Neural Networks (CNNs) have been adopted for texture
classification as well.  CNNs allow to leverage very large datasets of
labeled images, by learning intermediate image representations that
can be used for various image classification
problems~\cite{razavian2014cnn}.  For instance, Cimpoi et
al.~\cite{cimpoi2015deep} used Fisher Vectors to pool features
computed by a CNN trained for object recognition.

In addition to these general purpose texture descriptors, a variety of
descriptors have been specially designed to be robust with respect to
specific variations in the acquisition conditions.  Khan~\emph{et
  al.}~\cite{khan2010classical}, for instance, considered a
diagonal/offset model for illumination variations, deduced from it an
image normalization transformation, and finally extracted Gabor
features from the normalized images.  Other color normalization
techniques can be used for this purpose. Finlayson~\emph{et al.}
proposed rank-based features obtained from invariant color
representations~\cite{finlayson2005illuminant}.  Seifi~\emph{et al.},
instead, proposed to characterize color textures by analyzing the rank
correlation between pixels located in the same neighborhood. They
obtained a correlation measure which is related to the colors of the
pixels, and is not sensitive to illumination
changes~\cite{seifi2010color}.  Cusano et
al.~\cite{cusano2014combining} proposed a texture descriptor specially
designed to deal with the case of variations in the color of the
illuminant.  The reader can refer to the work of Drbohlav \emph{et
  al.}~\cite{drbohlav2010towards} for a comparative analysis of
texture methods under varying viewpoint and illumination, and to the
work of Kandaswamy \emph{et al.}~\cite{kandaswamy2011} for a
comparison among texture analysis schemes under non-ideal conditions.

In this work we compared several descriptors from the state of the
art, by taking a few representative descriptors for each of the
approaches mentioned above.
Several descriptors have been applied to both color and gray-scale
images, where the gray-scale image is defined as the luminance of the
image and is obtained by using the standard formula:
$L = 0.299R + 0.587G + 0.114B$.

In order to make the results readable we consider, here, only a
selection of all the descriptors evaluated.

\subsection{Traditional descriptors}

\begin{itemize}
\item 256-dimensional gray-scale histogram;
\item 512-dimensional Hue and Value marginal histogram obtained from
  the HSV color representation of the image;
\item 768-dimensional RGB and \emph{rgb} marginal
  histograms~\cite{Pietikainen1996};
\item 10-dimensional feature vector composed of normalized
  \emph{chromaticity moments} as defined in~\cite{Paschos2000};
\item 15-dimensional feature vector composed of contrast, correlation,
  energy, entropy and homogeneity extracted from the
  \emph{co-occurrence matrices} of each color
  channel~\cite{Arvis2004,Parkkinen1996};
\item 144-dimensional \emph{Gabor} features composed of mean and
  standard deviation of six orientations extracted at four frequencies
  for each color channel~\cite{Bianconi2011,Bianconi2007};
\item 264-dimensional \emph{opponent Gabor} feature vector extracted
  as Gabor features from several inter/intra channel combinations:
  monochrome features extracted from each channel separately and
  opponent features extracted from couple of colors at different
  frequencies~\cite{Jain1998};
\item 54-dimensional \emph{Dual Tree Complex Wavelet Transform}
  (DT-CWT) features obtained considering four scales, mean and
  standard deviation, and three color
  channels~\cite{Bianconi2011,Barilla2008};
\item 26-dimensional feature vector obtained calculating morphological
  operators (\emph{granulometries}) at four angles and for each color
  channel~\cite{Kandaswamy2005};
\item 512-dimensional \emph{Gist} features obtained considering eight
  orientations and four scales for each
  channel~\cite{oliva2001modeling};
\item 81-dimensional \emph{Histogram of Oriented Gradients} (HOG)
  feature vector~\cite{junior2009trainable}. Nine histograms with nine
  bins are concatenated to achieve the final feature vector;
\item 243-dimensional \emph{Local Binary Patterns} (LBP) feature
  vector computed with 16 neighbors, radius two and uniform patterns.
  We applied LBP to gray-scale images and then also to the color
  channels RGB, CIE-Lab and Ohta's \ohta{} spaces (in these cases the
  vector will be 729-dimensional)~\cite{maenpaa2004classification};
\item Combination of LBP computed on pairs of color channels, namely
  the \emph{Opponent Color LBP} (OCLBP)~\cite{chan2007multispectral};
\item LBP combined with the Local Color Contrast descriptor, as
  described in~\cite{cusano2014combining};
\item 499-dimensional Local Color Contrast feature vector. It is
  obtained by concatenating the LBP on the gray images with a
  quantized measure of color contrast~\cite{cusano2014combining};
\end{itemize}

\subsection{Descriptors for object recognition}

The features considered here consists in the aggregation of local
descriptors according to the quantization defined by a codebook of
visual words.  As local descriptors we used 128-dimensional dense SIFT
obtained from the gray-scale image by considering a spatial histogram
of local gradient orientations. The spatial bins have an extent of
\num{6 x 6}. The descriptors have been sampled every two pixel and at
scales $2^{i/3}$, $i=0,1,2,\dots$.

The object recognition features differ for the aggregation method, but
all of them are based on a codebook of \num{1024} visual words built
on images from external sources. In particular we downloaded
\num{20000} images from Flickr containing various content, such as
sunset, countryside, etc.\ and we used $k$-means to find \num{1024}
representative vectors.

The object recognition features considered here are:
\begin{itemize}
\item \num{1024}-dimensional bag of visual words (BoVW).
\item \num{25600}-dimensional vector of locally aggregated descriptors
  (vlad)~\cite{cimpoi2014describing}.
\item \num{40960}-dimensional Fisher's vectors (fv) of locally
  aggregated descriptors~\cite{jegou2010aggregating}.
\end{itemize}

\subsection{CNN-based descriptors}

The CNN-based features have been obtained as the intermediate
representations of deep convolutional neural networks originally
trained for object recognition. The networks are used to generate a
texture descriptor by removing the final softmax nonlinearity and the
last fully-connected layer, resulting in feature vectors which are
$L^{2}$ normalized before being used for classification. We considered
the most representative CNN architectures in the state of the
art~\cite{vedaldiCNN}, each exploring a different accuracy/speed
trade-off. All the CNNs have been trained on the ILSVRC-2012 dataset
using the same protocol as in~\cite{krizhevsky2012imagenet}. In
particular we considered \num{4096}, \num{2048}, \num{1024} and
\num{128}-dimensional feature vectors as
follows~\cite{razavian2014cnn}:
\begin{itemize}
\item \emph{BVLC AlexNet} (BVLC AlexNet): AlexNet trained on ILSVRC
  2012~\cite{krizhevsky2012imagenet}.
\item \emph{BVLC Reference CaffeNet} (BVLC Ref): AlexNet trained on
  ILSVRC 2012, with a minor variation from the version as described
  in~\cite{krizhevsky2012imagenet}.
\item \emph{Fast CNN} (Vgg F): it is similar to the one presented
  in~\cite{krizhevsky2012imagenet} with a reduced number of
  convolutional layers and the dense connectivity between
  convolutional layers. The last fully-connected layer is
  4096-dimensional~\cite{chatfield2014return}.
\item \emph{Medium CNN} (Vgg M): it is similar to the one presented
  in~\cite{zeiler2014visualizing} with a reduced number of filters in
  the convolutional layer four. The last fully-connected layer is
  4096-dimensional~\cite{chatfield2014return}.
\item \emph{Medium CNN} (Vgg M-2048-1024-128): three modifications of
  the Vgg M network, with lower dimensional last fully-connected
  layer. In particular we used a feature vector of 2048, 1024 and 128
  size~\cite{chatfield2014return}.
\item \emph{Slow CNN} (Vgg S): it is similar to the one presented
  in~\cite{sermanet2013overfeat} with a reduced number of
  convolutional layers, less filters in the layer five and the Local
  Response Normalization. The last fully-connected layer is
  4096-dimensional~\cite{chatfield2014return}.
\item \emph{Vgg Very Deep 19 and 16 layers} (Vgg VeryDeep 16 and 19):
  the configuration of these networks has been achieved by increasing
  the depth to 16 and 19 layers, that results in a substantially
  deeper network than what has been used in previous
  studies~\cite{simonyan2014very}.
\end{itemize}

\begin{figure*}[tb]
  \centering
  \begin{tabular}{cc}
    \includegraphics[width=0.46\linewidth]{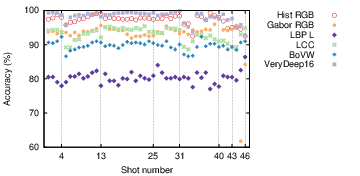} &
    \includegraphics[width=0.46\linewidth]{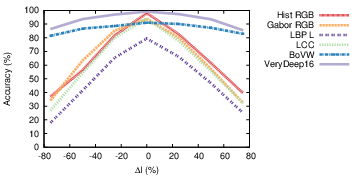} \\
    (a)&(b)\\
    \includegraphics[width=0.46\linewidth]{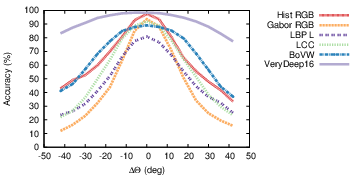} &
    \includegraphics[width=0.46\linewidth]{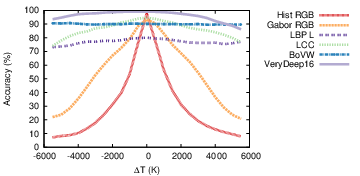} \\
    (c)&(d)\\
  \end{tabular}
  \caption{Detail of the classification rates as functions of the
    amount of variability in the illumination conditions between the
    training and test set. (a) Accuracy obtained in the \emph{no
      variations} classification task (each point corresponds to one
    of the 46 shots).  (b) Accuracy with respect to the difference
    $\Delta I$ of \emph{light intensity}. (c) Accuracy obtained
    varying the difference between the direction of the light. (d)
    Accuracy with respect to the difference $\Delta T$ of
    \emph{Daylight temperature}.}
  \label{fig:behavior}
\end{figure*}

\subsection{Color normalization}

Invariance with respect to specific changes in acquisition conditions,
such as those caused by variations in the illumination, is an
important property of visual descriptors.  Illumination variations can
be also compensated by preprocessing images with a color normalization
method. Color normalization methods try to assign a constant color to
objects acquired under different illumination conditions.

In order to evaluate this strategy, we have preprocessed the \raw{}
database by using several existing normalization methods and next we
have extracted features by using the best color descriptors from the
set of descriptors evaluated in table~\ref{tab:descriptors_res}. More
precisely, we considered two implementations of the Retinex method
described in \cite{funt2000retinex} that improve the computational
efficiency while preserving the underlying principles: the
\emph{McCann99}~\cite{mccann1999lessons} and the
\emph{Frankle-McCann}~\cite{frankle1983method}. Furthermore, we
considered the Gray World \cite{finlayson2004shades}, two variants of
edge based algorithm, the Gray-Edge \cite{van2007edge-based} and the
weighted Gray-Edge method \cite{gijsenij2012improving}.

\section{Experiments}
\label{sec:experiments}

In all the experiments we used the nearest neighbor classification
strategy: given a patch in the test set, its distance with respect to
all the training patches is computed. The prediction of the classifier
is the class of the closest element in the training set. For this
purpose, after some preliminary tests with several descriptors in
which we evaluated the most common distance measures, we decided to
use the $L1$ distance:
$d(\bf{x},\bf{y})=\sum_{i=1}^{N} | x_i - y_i |$, where $\bf{x}$ and
$\bf{y}$ are two feature vectors.  All the experiments have been
conducted under the \emph{maximum ignorance} assumption, that is, no
information about the lighting conditions of the test patches is
available for the classification method and for the
descriptors. Performance is reported as classification rate (i.e., the
ratio between the number of correctly classified images and the number
of test images).  Note that more complex classification schemes
(e.g.~SVMs) would have been viable.  We decided to adopt the simplest
one in order to focus the evaluation on the features themselves and
not on the classifier.

\subsection{\raw{} database setup}

For each of the 68 classes we considered 16 patches obtained by
dividing the original texture image, that is of size \num{800 x 800}
pixels, in 16 non-overlapping squares of size \num{200 x 200}
pixels. For each class we selected eight patches for training and
eight for testing alternating them in a chessboard pattern.  We form
subsets of $68 \times (8 + 8) = 1088$ patches by taking the training
and test patches from images taken under different lighting
conditions.

\begin{figure*}[tb]
\tiny
\centering
\setlength{\tabcolsep}{1pt}
  \begin{tabular}{cccc}
    \includegraphics[width=0.247\linewidth]{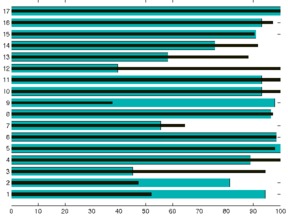} &
    \includegraphics[width=0.247\linewidth]{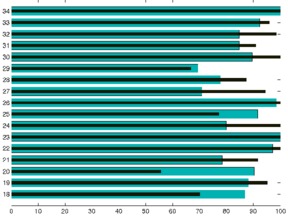} &
    \includegraphics[width=0.247\linewidth]{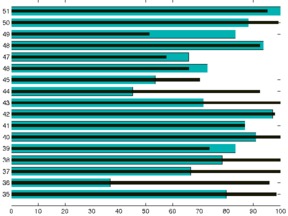} &
    \includegraphics[width=0.247\linewidth]{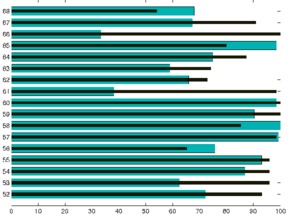} \\
    	\multicolumn{4}{c}{(a)}\\[12pt]
    \includegraphics[width=0.247\linewidth]{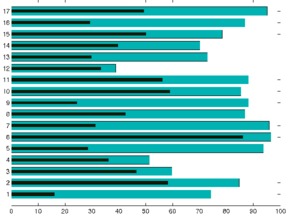} &
    \includegraphics[width=0.247\linewidth]{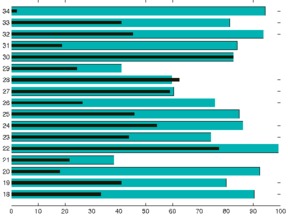} &
    \includegraphics[width=0.247\linewidth]{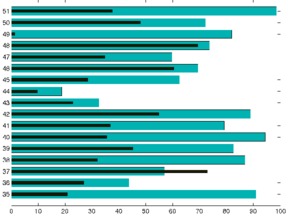} &
    \includegraphics[width=0.247\linewidth]{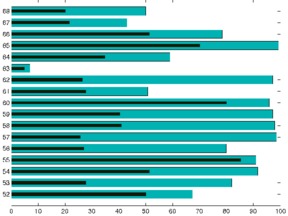} \\
    	\multicolumn{4}{c}{(b)}\\
    \end{tabular}
    \caption{Accuracy of the \emph{Vgg VeryDeep 16} (turquoise) and
      \emph{BoVW} (black) over the 68 classes. (a) training and test
      images are under different lights and same angle. (b) training
      and test images are under the same light and different angles.
      To map the numbers to the corresponding classes see
      Figure~\ref{fig:db_overview}.}
  \label{fig:same_light_different_angle}
\end{figure*}

\begin{table*}[tb]
  \caption{Classification rates (\%) of the texture descriptors
    considered. For each classification task, the best result is
    reported in bold. }
  \scriptsize
  \setlength{\tabcolsep}{3.5pt}
  \centering
  \begin{tabular}{lccccccccc}
    \toprule
Features	&	No variations &Light intensity	&Light direction&	Daylight temp. 	&	LED temperature	& Daylight vs.~LED & 	Temp.~OR Dir. & 	Temp.~\& Dir.	& Multi-illum \\
	&	avg (min) &avg (min)& avg (min)	&avg (min)	& avg (min) &avg (min)&avg (min)	& avg (min) &avg (min)	\\
 \midrule
Hist. L	&	78.32 (60.66)&	6.77 (1.47)&	20.97 (2.21)&	49.94 (11.95)&	27.18 (5.88)&	38.05 (6.43)&	10.45 (1.29)&	7.98 (1.29)&	47.33 (30.15)\\
Hist. H V 	&	96.38 (84.56)&	31.45 (14.52)&	51.01 (13.97)&	49.11 (9.93)&	51.56 (23.35)&	44.39 (9.19)&	16.47 (4.23)&	11.36 (4.23)&	49.82 (38.42)\\
Hist. RGB	&	94.93 (87.13)&	15.89 (3.12)&	40.82 (5.70)&	56.45 (18.20)&	37.51 (12.68)&	43.44 (8.00)&	15.53 (2.76)&	10.21 (2.76)&	53.83 (42.28)\\
Hist. \emph{rgb}	&	97.24 (92.46)&	67.08 (36.95)&	64.07 (24.63)&	37.35 (6.43)&	17.38 (3.31)&	25.71 (5.15)&	20.16 (2.39)&	12.57 (2.39)&	45.25 (16.36)\\[1.5ex]
Chrom. mom.	&	82.54 (58.46)&	68.43 (48.90)&	53.68 (22.24)&	33.41 (4.96)&	18.66 (3.68)&	24.16 (5.06)&	17.03 (2.21)&	10.63 (2.21)&	33.92 (20.77)\\
Coocc. matr.	&	35.33 (9.93)&	7.20 (2.02)&	9.39 (0.55)&	23.02 (9.74)&	19.01 (6.62)&	19.88 (5.61)&	3.30 (0.18)&	2.89 (0.18)&	10.48 (6.07)\\
Coocc. matr. L	&	18.68 (1.47)&	3.32 (0.00)&	5.63 (0.55)&	16.99 (6.99)&	9.49 (3.31)&	12.94 (2.85)&	2.49 (0.00)&	2.57 (0.00)&	6.86 (2.57)\\
DT-CWT	&	92.26 (81.62)&	21.68 (1.65)&	49.48 (12.32)&	66.29 (25.92)&	42.31 (14.34)&	49.77 (15.44)&	19.23 (3.12)&	13.22 (3.12)&	65.93 (55.33)\\
DT-CWT L	&	72.85 (58.09)&	10.65 (1.29)&	29.29 (5.15)&	60.13 (27.39)&	32.70 (4.04)&	44.06 (5.06)&	14.70 (1.47)&	12.92 (1.47)&	52.73 (37.13)\\[1.5ex]
Gabor RGB	&	93.02 (61.76)&	66.96 (32.35)&	51.66 (11.95)&	64.81 (20.77)&	38.13 (12.13)&	48.03 (12.59)&	27.18 (3.49)&	16.43 (3.49)&	75.18 (59.93)\\
Gabor L	&	72.91 (70.04)&	46.57 (18.75)&	29.34 (3.86)&	68.94 (59.56)&	67.62 (58.82)&	66.86 (53.40)&	27.58 (2.57)&	16.09 (2.57)&	74.33 (72.79)\\
Opp. Gabor RGB	&	96.15 (59.38)&	21.51 (3.49)&	50.61 (12.13)&	67.75 (22.98)&	41.78 (14.34)&	50.80 (15.07)&	20.22 (3.86)&	13.47 (3.86)&	58.58 (43.01)\\
Gist RGB	&	66.20 (62.50)&	55.06 (31.99)&	45.75 (10.85)&	55.49 (28.31)&	36.78 (13.24)&	43.41 (13.79)&	25.13 (2.76)&	17.30 (2.76)&	60.36 (54.78)\\
Granulometry	&	91.98 (51.65)&	63.73 (27.76)&	48.65 (10.66)&	69.80 (21.51)&	33.58 (6.80)&	48.79 (6.34)&	22.20 (1.65)&	12.86 (1.65)&	74.23 (64.34)\\
HoG	&	46.74 (43.20)&	37.52 (24.82)&	27.48 (8.46)&	41.14 (29.60)&	35.29 (22.24)&	36.30 (19.30)&	16.99 (3.49)&	11.71 (3.49)&	43.66 (40.62)\\[1.5ex]
LBP L	&	80.37 (77.02)&	51.15 (17.83)&	47.09 (9.19)&	77.76 (72.24)&	70.77 (54.60)&	73.15 (55.06)&	29.54 (5.51)&	18.66 (5.51)&	76.99 (74.26)\\
LBP RGB	&	93.55 (90.81)&	68.87 (33.46)&	59.48 (13.60)&	72.40 (24.63)&	48.39 (15.07)&	56.08 (16.82)&	23.72 (0.55)&	14.19 (0.55)&	76.81 (67.10)\\
LBP Lab	&	92.90 (88.42)&	71.88 (32.54)&	61.74 (17.28)&	70.61 (24.08)&	51.53 (21.69)&	56.00 (19.49)&	27.55 (3.31)&	18.06 (3.31)&	77.21 (71.32)\\
LBP $\text{I}_1\text{I}_2\text{I}_3$	&	91.40 (82.90)&	66.28 (28.12)&	63.05 (17.10)&	70.58 (25.92)&	49.90 (18.38)&	54.76 (17.00)&	27.05 (1.10)&	18.39 (1.10)&	75.55 (64.89)\\
OCLBP	&	95.92 (92.28)&	78.75 (51.47)&	65.70 (14.52)&	67.92 (19.67)&	49.94 (15.81)&	53.93 (15.81)&	25.73 (1.65)&	16.99 (1.65)&	76.53 (50.00)\\
LCC	&	92.92 (88.60)&	62.64 (26.84)&	56.15 (12.13)&	88.78 (73.71)&	74.25 (46.88)&	78.82 (50.64)&	31.13 (5.15)&	19.85 (5.15)&	85.91 (84.38)\\[1.5ex]
BoVW	&	89.73 (86.58)&	87.38 (81.43)&	67.38 (16.18)&	90.02 (88.05)&	88.87 (86.76)&	89.53 (87.68)&	51.59 (12.68)&	39.34 (12.68)&	88.60 (87.68)\\
VLAD	&	79.29 (75.18)&	76.87 (70.22)&	64.73 (20.77)&	78.70 (76.65)&	77.87 (73.90)&	78.31 (75.74)&	51.44 (18.20)&	42.62 (18.75)&	80.58 (79.78)\\
FV	&	85.59 (80.51)&	81.02 (71.32)&	69.31 (27.39)&	86.57 (84.38)&	84.58 (78.68)&	85.51 (82.35)&	55.26 (20.04)&	45.95 (20.04)&	85.69 (84.56)\\[1.5ex]
Vgg F	&	96.94 (89.34)&	88.07 (72.61)&	83.23 (47.06)&	96.27 (86.95)&	87.94 (65.26)&	89.40 (62.96)&	54.89 (13.42)&	45.62 (13.42)&	95.59 (94.49)\\
Vgg M	&	97.50 (89.89)&	89.94 (76.29)&	86.14 (52.02)&	97.30 (88.97)&	90.64 (69.85)&	91.91 (71.60)&	60.25 (20.04)&	51.13 (20.04)&	96.23 (94.49)\\
Vgg M S	&	97.53 (91.36)&	90.93 (74.82)&	86.24 (51.10)&	96.16 (85.66)&	89.90 (68.75)&	90.33 (69.21)&	59.02 (16.73)&	49.48 (16.73)&	95.53 (93.57)\\
Vgg M 2048	&	97.27 (89.89)&	89.08 (72.43)&	84.74 (48.35)&	97.38 (90.81)&	90.06 (64.34)&	91.41 (69.58)&	57.18 (14.15)&	47.89 (14.15)&	95.01 (93.38)\\
Vgg M 1024	&	96.79 (87.87)&	88.53 (69.67)&	84.15 (49.08)&	96.90 (90.07)&	90.14 (63.42)&	91.47 (69.76)&	56.26 (14.15)&	47.26 (14.15)&	93.75 (91.91)\\
Vgg M 128	&	93.50 (78.31)&	81.54 (59.93)&	78.04 (45.04)&	93.01 (82.35)&	79.72 (46.69)&	83.20 (50.92)&	49.13 (13.79)&	40.75 (13.79)&	89.00 (87.50)\\
BVLC Ref	&	95.22 (84.56)&	83.23 (60.66)&	79.00 (40.81)&	94.51 (80.33)&	85.11 (58.64)&	86.52 (58.00)&	49.50 (9.38)&	40.63 (9.38)&	91.24 (89.15)\\
BVLC AlexNet	&	94.88 (82.17)&	85.00 (65.07)&	80.25 (43.57)&	95.79 (86.03)&	88.56 (66.36)&	89.64 (65.99)&	48.08 (9.38)&	39.37 (9.38)&	90.44 (88.24)\\
Vgg VeryDeep 16	&	\textbf{98.21} (92.83)&	\textbf{94.10} (85.48)&	\textbf{91.23} (66.73)&	\textbf{97.41} (86.21)&	93.69 (77.57)&	\textbf{93.67} (75.83)&	\textbf{70.81} (33.82)&	\textbf{63.64} (33.82)&	\textbf{96.60} (94.85)\\
Vgg VeryDeep 19	&	97.69 (92.10)&	93.01 (82.90)&	90.32 (62.50)&	96.98 (85.85)&	\textbf{93.94} (79.23)&	93.53 (74.82)&	69.64 (37.68)&	62.35 (37.68)&	95.59 (94.49)\\
   \bottomrule
  \end{tabular}
  \label{tab:descriptors_res}
\end{table*}

In this way we defined several subsets, grouped in nine texture
classification tasks.
\begin{enumerate}
\item \textbf{No variations}: 46 subsets. Each subset is composed of
  training and test patches taken under the same lighting condition.

\item \textbf{Light intensity}: 12 subsets obtained by combining the
  four intensity variations. Each subset is composed of training and
  test patches with different light intensity values.

\item \textbf{Light direction}: 72 subsets obtained by combining the
  nine different light directions. Each subset is composed of training
  and test patches with different light direction.

\item \textbf{Daylight temperature}: 132 subsets obtained by combining
  all the 12 daylight temperature variations. Each subset is composed
  of training and test patches with different light temperatures.

\item \textbf{LED temperature}: 30 subsets obtained by combining all
  the six LED temperature variations. Each subset is composed of
  training and test patches with different light temperatures.

\item \textbf{Daylight vs. LED}: 72 subsets obtained by combining 12
  daylight temperatures with six LED temperatures.

\item \textbf{Temperature or direction}: 72 subsets obtained by
  combining all the nine combinations of color temperatures and light
  directions. Each subset is composed of training and test patches
  where either the color or the direction (or both) change.

\item \textbf{Temperature and direction}: 36 subsets obtained by
  combining all the nine combinations of color temperatures and light
  directions. Each subset is composed of training and test patches
  where both the color and the direction change.

\item \textbf{Multiple illuminant}: six subsets obtained by combining
  the three acquisitions with multiple illuminants.
\end{enumerate}

\begin{table*}[tb]
  \caption{Classification rates (\%)  of a selection of color
    descriptors combined with different preprocessing methods. }
  \scriptsize
  \setlength{\tabcolsep}{2.5pt}
  \centering
  \begin{tabular}{lccccccccc}
    \toprule
    Features	&	No variations &Light intensity	&Light direction&	Daylight temp. 	&	LED temperature	& Daylight vs.~LED & 	Temp. OR Dir. & 	Temp. \& Dir.	& Multi-illum \\
&	avg (min) &avg (min)& avg (min)	&avg (min)	& avg (min) &avg (min)&avg (min)	& avg (min) &avg (min)	\\
 \midrule
 \textbf{VGG VeryDeep 16}	&	98.21 (92.83)&	94.10 (85.48)&	91.23 (66.73)&	97.41 (86.21)&	93.69 (77.57)&	93.67 (75.83)&	70.81 (33.82)&	63.64 (33.82)&	96.60 (94.85)\\[0.5ex]
Retinex McCann	&	98.84 (94.67)&	94.04 (86.40)&	92.83 (67.65)&	96.72 (80.88)&	93.43 (78.68)&	93.04 (77.11)&	72.47 (33.46)&	65.70 (33.46)&	96.97 (96.14)\\
Retinex Frankle	&	98.91 (94.49)&	94.47 (87.32)&	93.49 (70.22)&	97.13 (83.46)&	94.20 (79.41)&	93.96 (79.87)&	75.33 (40.81)&	68.96 (40.81)&	97.40 (96.32)\\
Gray-World	&	98.24 (87.32)&	98.35 (95.59)&	91.30 (62.13)&	96.26 (83.09)&	78.20 (46.51)&	83.57 (45.96)&	59.85 (24.26)&	50.20 (24.26)&	96.17 (94.49)\\
Gray-Edge	&	98.64 (89.52)&	90.87 (81.43)&	90.93 (66.18)&	97.56 (92.28)&	93.21 (81.43)&	94.09 (81.43)&	73.82 (41.18)&	66.76 (41.18)&	96.78 (95.77)\\
Weighted Gray-Edge	&	98.37 (85.11)&	91.18 (78.31)&	90.69 (65.26)&	97.62 (93.01)&	93.24 (81.99)&	94.18 (81.16)&	73.57 (41.36)&	66.45 (41.36)&	96.51 (94.49)\\[1.5ex]
 \textbf{LBP RGB}	&	93.55 (90.81)&	68.87 (33.46)&	59.48 (13.60)&	72.40 (24.63)&	48.39 (15.07)&	56.08 (16.82)&	23.72 (0.55)&	14.19 (0.55)&	76.81 (67.10)\\[0.5ex]
Retinex McCann	&	94.07 (91.18)&	69.61 (38.24)&	60.77 (13.60)&	76.21 (31.80)&	56.59 (26.47)&	61.71 (23.71)&	28.42 (2.94)&	18.16 (2.94)&	82.69 (76.47)\\
Retinex Frankle	&	94.21 (90.62)&	68.37 (33.64)&	58.37 (14.71)&	71.99 (24.45)&	47.73 (16.54)&	55.40 (16.54)&	24.71 (2.21)&	14.92 (2.21)&	75.46 (64.71)\\
Gray-World	&	93.63 (90.81)&	80.91 (62.68)&	61.94 (13.42)&	77.88 (37.68)&	47.09 (12.68)&	58.19 (13.51)&	27.10 (0.74)&	17.11 (0.74)&	72.40 (63.05)\\
Gray-Edge	&	94.03 (91.18)&	62.96 (27.02)&	58.66 (14.15)&	72.79 (30.33)&	44.01 (10.66)&	54.02 (13.60)&	22.75 (0.37)&	13.30 (0.37)&	75.80 (65.99)\\
Weighted Gray-Edge	&	93.93 (81.62)&	63.45 (26.65)&	58.87 (13.97)&	72.83 (27.76)&	44.09 (10.85)&	53.98 (14.25)&	22.95 (0.55)&	13.40 (0.55)&	76.29 (67.28)\\[1.5ex]
 \textbf{Hist \emph{rgb}}	&	97.24 (92.46)&	67.08 (36.95)&	64.07 (24.63)&	37.35 (6.43)&	17.38 (3.31)&	25.71 (5.15)&	20.16 (2.39)&	12.57 (2.39)&	45.25 (16.36)\\[0.5ex]
Retinex McCann	&	98.66 (95.77)&	65.40 (38.60)&	57.01 (18.93)&	32.79 (7.17)&	17.54 (2.76)&	23.76 (5.70)&	16.22 (2.21)&	10.06 (2.39)&	38.91 (28.49)\\
Retinex Frankle	&	98.82 (95.77)&	66.42 (39.52)&	57.47 (19.67)&	34.81 (6.99)&	17.95 (3.12)&	24.45 (5.24)&	16.73 (2.76)&	10.15 (2.76)&	41.42 (28.49)\\
Gray-World	&	98.81 (96.32)&	47.87 (19.30)&	36.08 (5.70)&	51.90 (10.48)&	22.42 (1.10)&	35.12 (0.46)&	13.41 (0.00)&	8.36 (0.00)&	27.24 (10.66)\\
Gray-Edge	&	98.36 (96.32)&	78.80 (59.38)&	70.90 (31.80)&	64.46 (17.10)&	37.95 (8.46)&	46.55 (9.47)&	33.42 (6.25)&	24.51 (6.25)&	58.92 (35.29)\\
Weighted Gray-Edge	&	98.16 (84.01)&	75.97 (55.15)&	69.37 (29.60)&	64.62 (18.38)&	38.65 (9.74)&	46.90 (8.46)&	33.08 (6.99)&	24.59 (6.99)&	52.30 (30.88)\\[1.5ex]
 \textbf{Gabor RGB}	&	93.02 (61.76)&	66.96 (32.35)&	51.66 (11.95)&	64.81 (20.77)&	38.13 (12.13)&	48.03 (12.59)&	27.18 (3.49)&	16.43 (3.49)&	75.18 (59.93)\\[0.5ex]
Retinex McCann	&	92.91 (89.34)&	66.88 (32.90)&	51.08 (12.68)&	65.87 (22.24)&	37.55 (10.66)&	48.24 (11.40)&	22.96 (0.92)&	13.12 (0.92)&	75.80 (64.71)\\
Retinex Frankle	&	93.57 (90.07)&	67.31 (33.27)&	51.51 (12.68)&	66.02 (21.88)&	37.71 (11.03)&	48.50 (11.95)&	23.24 (0.92)&	13.15 (0.92)&	75.77 (64.15)\\
Gray-World	&	94.06 (92.28)&	70.54 (37.32)&	52.43 (13.42)&	68.40 (23.35)&	40.29 (11.76)&	50.77 (12.68)&	26.27 (2.02)&	15.18 (2.02)&	78.40 (67.46)\\
Gray-Edge	&	93.46 (60.85)&	64.57 (31.43)&	51.51 (11.95)&	66.05 (22.79)&	38.92 (11.95)&	48.78 (12.68)&	27.08 (3.31)&	16.43 (3.31)&	75.06 (59.56)\\
Weighted Gray-Edge	&	93.40 (62.13)&	64.41 (31.25)&	51.50 (11.76)&	65.90 (23.16)&	38.74 (12.13)&	48.66 (13.05)&	27.00 (3.49)&	16.37 (3.49)&	75.03 (60.85)\\[1.5ex]
 \textbf{LCC}	&	92.92 (88.60)&	62.64 (26.84)&	56.15 (12.13)&	88.78 (73.71)&	74.25 (46.88)&	78.82 (50.64)&	31.13 (5.15)&	19.85 (5.15)&	85.91 (84.38)\\[0.5ex]
Retinex McCann	&	92.41 (88.60)&	64.11 (27.57)&	56.97 (12.13)&	85.80 (63.60)&	63.88 (30.88)&	71.12 (31.99)&	29.90 (7.54)&	20.47 (7.54)&	85.63 (82.17)\\
Retinex Frankle	&	92.38 (87.68)&	62.33 (23.53)&	56.14 (12.87)&	84.63 (58.27)&	61.64 (26.65)&	69.27 (28.58)&	28.93 (5.88)&	19.26 (5.88)&	84.53 (81.43)\\
Gray-World	&	92.93 (87.50)&	74.39 (43.75)&	60.25 (15.07)&	75.50 (33.46)&	42.67 (8.27)&	54.36 (7.81)&	26.51 (3.86)&	17.02 (3.86)&	78.12 (71.88)\\
Gray-Edge	&	94.27 (89.15)&	57.41 (21.88)&	55.10 (10.11)&	82.16 (48.35)&	49.74 (11.76)&	61.45 (13.24)&	24.37 (2.94)&	16.40 (2.94)&	87.10 (85.11)\\
Weighted Gray-Edge	&	93.92 (81.99)&	59.19 (23.71)&	54.36 (11.03)&	82.46 (50.92)&	50.03 (11.03)&	61.52 (13.79)&	24.17 (2.39)&	16.20 (2.39)&	86.70 (84.74)\\
    \bottomrule
\end{tabular}
\label{tab:normalization_res}
\end{table*}

\section{Results}
\label{sec:results}

Table~\ref{tab:descriptors_res} reports the performance obtained by
the descriptors considered as \emph{average} and \emph{minimum
  accuracy} over the nine classification tasks.  For the four main
tasks (\emph{same illuminant}, \emph{light intensity}, \emph{light
  direction} and \emph{daylight temperature}) the results are shown in
greater detail in Figure~\ref{fig:behavior}, but only for some
representative methods.  When training and test are taken under the
same lighting conditions the classification rates are generally high,
regardless the specific conditions.  CNN features perform very well,
with a peak of 98.2\% of accuracy obtained with the features extracted
by the \emph{Vgg Very Deep 16} network.  Other, more traditional
features perform very well in this scenario (OCLBP at 95.9\%
Opp.~Gabor at 96.2\%\dots) and even simple $rgb$ histograms achieve an
accuracy of 97.2\%.  It is clear that under fixed conditions, texture
classification is not a very challenging problem, see also
Figure~\ref{fig:behavior}(a).

When training and test patches are taken under variable intensity, the
behavior of CNN features and of the descriptors taken from the object
recognition literature (\emph{BoVW}) is very stable.  Surprisingly,
traditional \emph{hand-crafted} features are heavily affected by this
kind of variations even when they are supposed to be robust to them,
as should be the case of LBP and Gabor-based features. This behavior
is more evident looking at Figure~\ref{fig:behavior}(b), where only
\emph{Vgg VeryDeep 16} and \emph{BoVW} have flat curves over changes
in the intensity of the light.

Perhaps one of the most challenging variation to take into account is
that related to the direction of the light, see Figure~\ref{fig:behavior}(c). In this task all the
descriptors suffered a noticeable decrease in performance.  However,
some CNN features remained, on average, above 90\% of accuracy.  The
performance of all the other features dropped below 70\%.

When the illuminant color is allowed to vary between train and test
images, the achromatic features are the least affected.  In
particular, the features from the object recognition literature
obtained the same performance of the \emph{same illumination} task.
For other features, such as LBP-L, we observed a decrease in the
performance, probably due to the variation in intensity caused by the
change of the color temperature.  Features that use color information
greatly suffer this kind of variability (see Figure~\ref{fig:behavior}(d)).  The most important exception is represented by the CNN features
that have been trained to exploit color, but in such a way to be
robust with respect to the large amount of variability in the object
categories they were supposed to discriminate.

Very low performance have been obtained when both direction and color
change simultaneously.  In this case the best results have been
obtained by, again, features from CNNs.  However, the highest
classification rate is quite low (about 63.6\%) and most networks do
not allow to achieve more than 50\%.  The results for the other
features are even worse than that. 

The last task involved the presence of multiple illuminants.  Since
their position was stable, we obtained similar results of those of the
case of variable color temperature.

Summing up, the challenges of recognizing textures under variable
illumination conditions greatly depends on the type of variability
involved in the experiments.  Features extracted by CNNs significantly
outperform the other descriptors considered.  Features from the object
recognition literature clearly outperform traditional
\emph{hand-crafted} texture features in all the scenarios considered.
Only under some specific circumstances these last ones outperformed
CNN features.  For instance, in Figure~\ref{fig:behavior} (d) it can
be observed that CNN features fall below the bag of visual words
descriptor for extreme variations in the color temperature. This
circumstances can be better understood by looking at
Fig.~\ref{fig:same_light_different_angle}(a). This figure compares the
behavior of \emph{Vgg Very Deep 16} and \emph{BoVW} over the 68
classes. Here the training and test images have been taken under
different lights but all directed with the same angle. In particular
we averaged the accuracy obtained in three sets of experiments, one
for each angle (24,60,90 deg) all including three lights: D65, D95 and
L27. It is quite evident that in this case object recognition features
outperform CNNs especially for those classes whose appearance is most
sensitive to color changes and that contain a more fine-grained
texture, such as \emph{(3) salt}, \emph{(12) sliced bread}, \emph{(36)
  flour}, \emph{(53) mango}, \emph{(61) coconut flakes}, \emph{(66)
  sugar}, etc. This result is due to the fact the CNNs have mainly
been trained on images of objects and thus containing more coarse
details.  In contrast, when the training and test images have been
taken under the same light but with different light directions, CNN
features demonstrate to be more robust than object recognition
features, see Fig.~\ref{fig:same_light_different_angle}(b). Here the
worst results are obtained by \emph{BoVW} on coarse-grained texture
images, such as \emph{(1) chickpeas}, \emph{(20) basmati rice},
\emph{(62) chicory}, etc.

\begin{figure*}[tb]
\tiny
\centering
\setlength{\tabcolsep}{1pt}
  \begin{tabular}{ccccc@{\hskip 0.5in}ccc}
   & \multicolumn{3}{c}{(06) candies}&&\multicolumn{3}{c}{(51) pink pepper}\\
    \cline{2-4}\cline{6-8}\\
    Original&
    \includegraphics[width=0.1\linewidth]{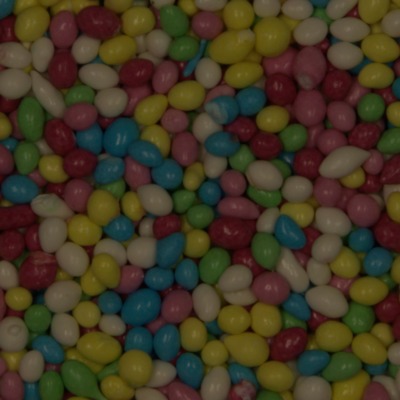} &
    \includegraphics[width=0.1\linewidth]{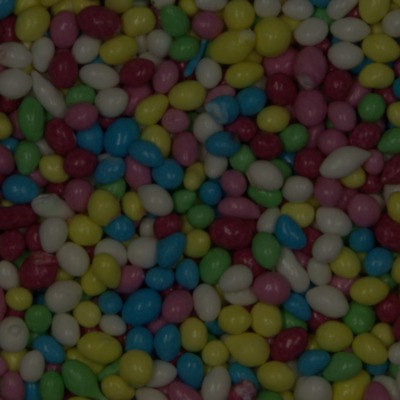} &
        \includegraphics[width=0.1\linewidth]{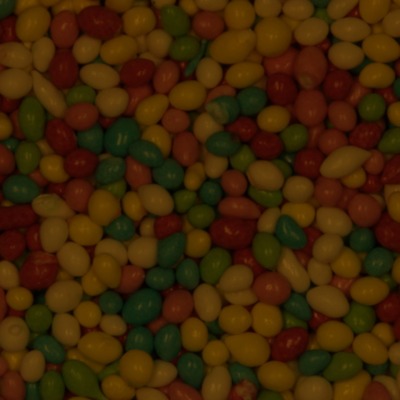} &&
 	 \includegraphics[width=0.1\linewidth]{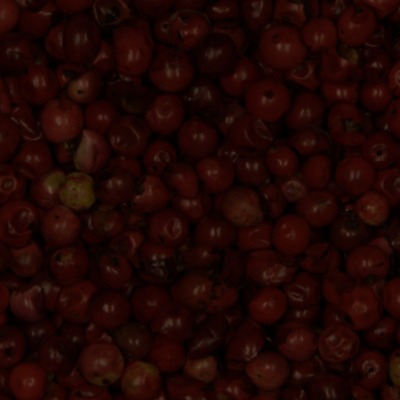} &
        \includegraphics[width=0.1\linewidth]{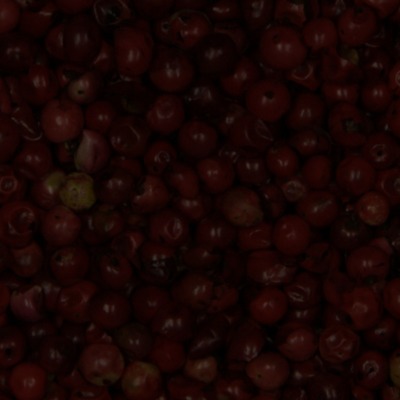} &
    \includegraphics[width=0.1\linewidth]{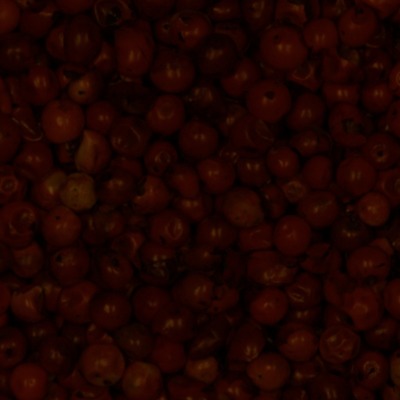} \\[4pt]
        Retinex McCann &
    \includegraphics[width=0.1\linewidth]{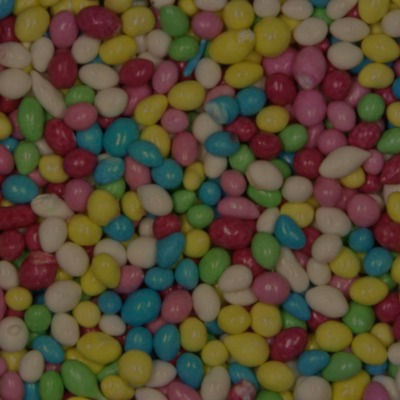} &
        \includegraphics[width=0.1\linewidth]{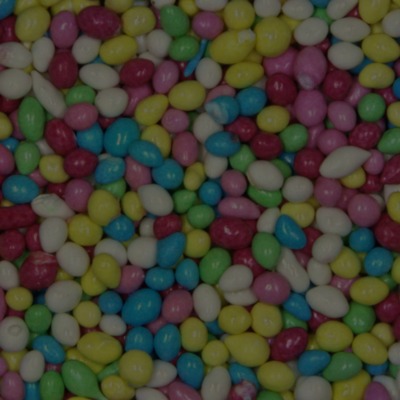} &
    \includegraphics[width=0.1\linewidth]{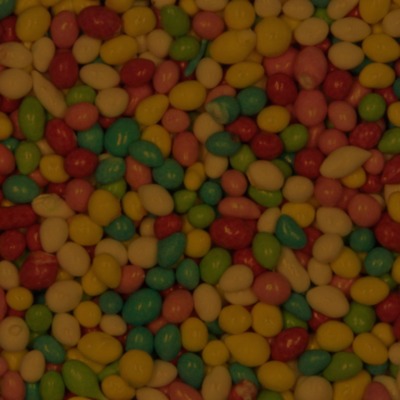} &&
  \includegraphics[width=0.1\linewidth]{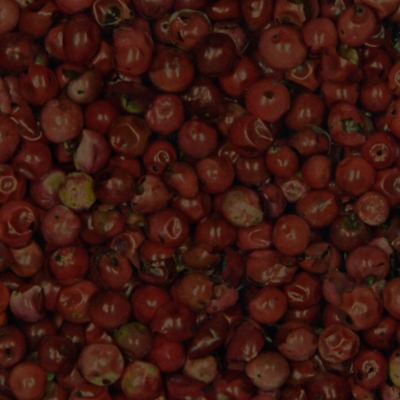} &
        \includegraphics[width=0.1\linewidth]{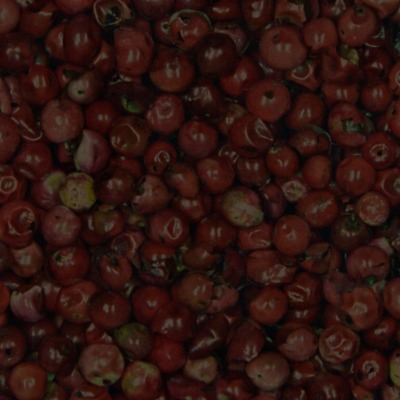} &
    \includegraphics[width=0.1\linewidth]{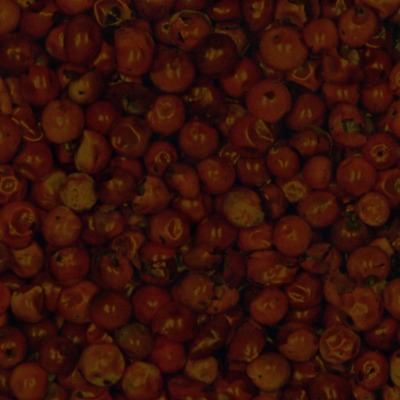} \\[4pt]
	Retinex Frankle &
        \includegraphics[width=0.1\linewidth]{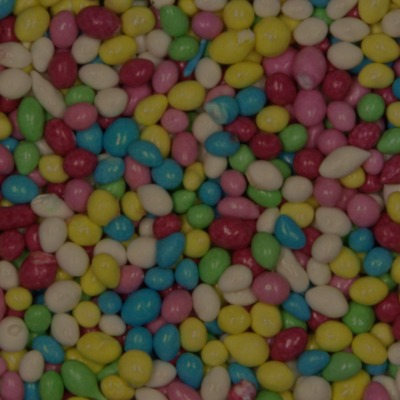} &
        \includegraphics[width=0.1\linewidth]{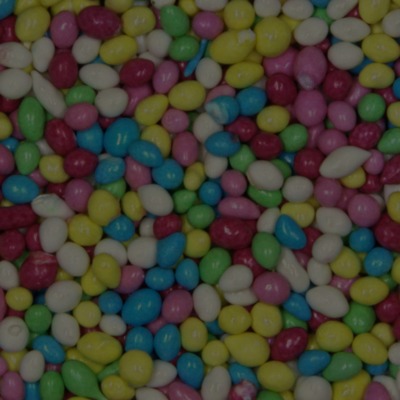} &
    \includegraphics[width=0.1\linewidth]{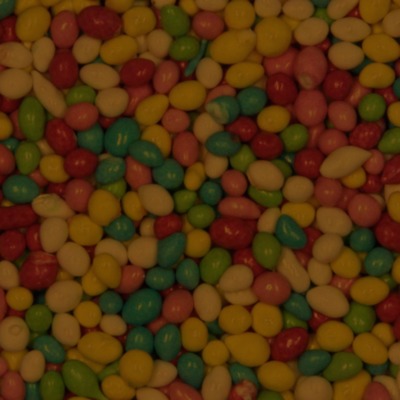} &&
  \includegraphics[width=0.1\linewidth]{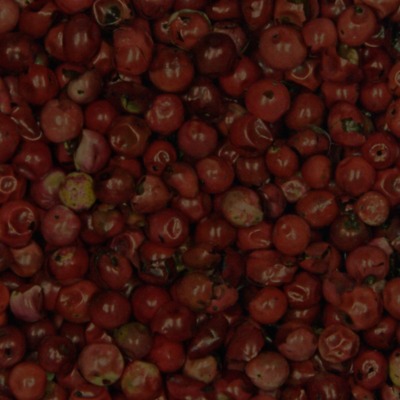} &
        \includegraphics[width=0.1\linewidth]{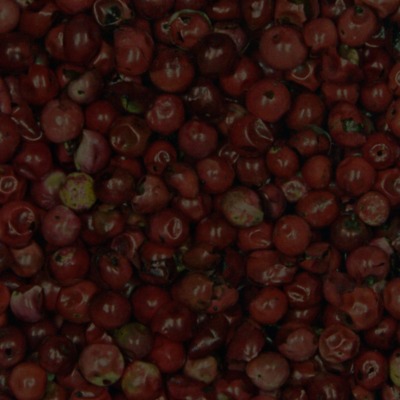} &
    \includegraphics[width=0.1\linewidth]{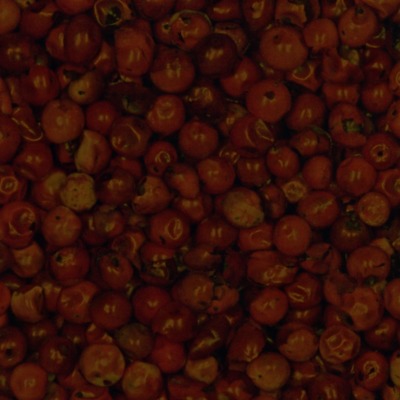} \\[4pt]
        Gray-World &
    \includegraphics[width=0.1\linewidth]{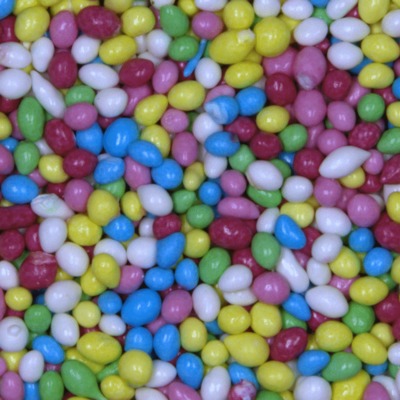} &
        \includegraphics[width=0.1\linewidth]{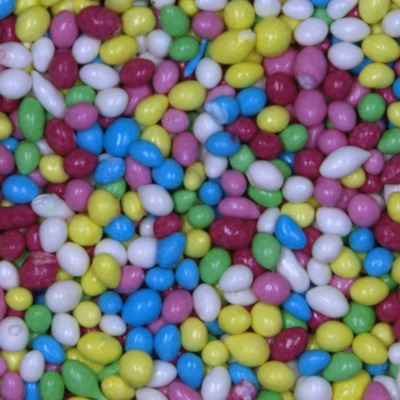} &
    \includegraphics[width=0.1\linewidth]{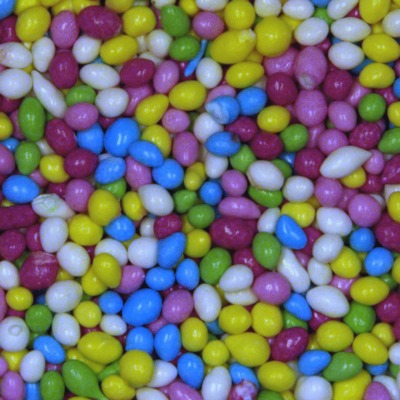} &&
  \includegraphics[width=0.1\linewidth]{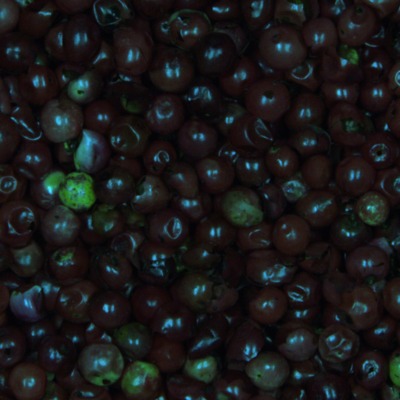} &
        \includegraphics[width=0.1\linewidth]{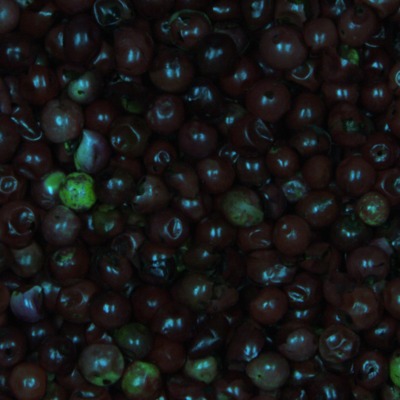} &
    \includegraphics[width=0.1\linewidth]{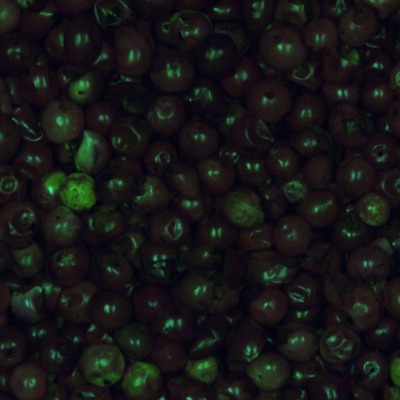} \\[4pt]
        Gray-Edge &
    \includegraphics[width=0.1\linewidth]{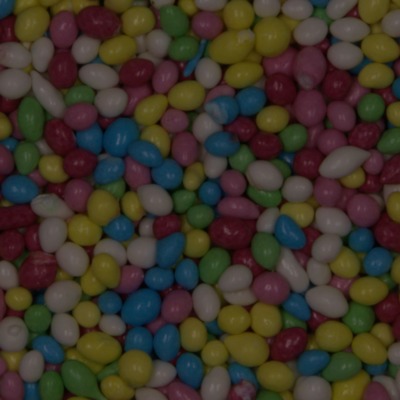} &
        \includegraphics[width=0.1\linewidth]{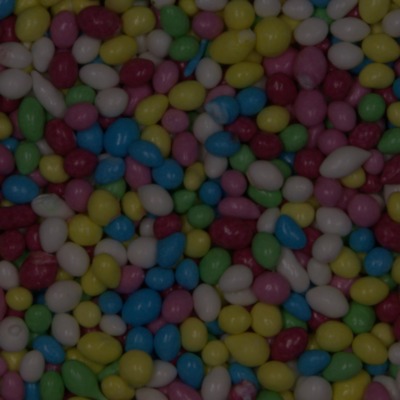} &
    \includegraphics[width=0.1\linewidth]{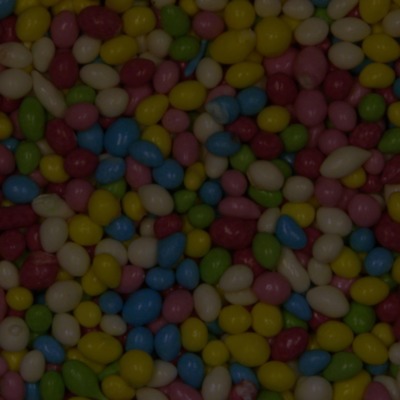} &&
  \includegraphics[width=0.1\linewidth]{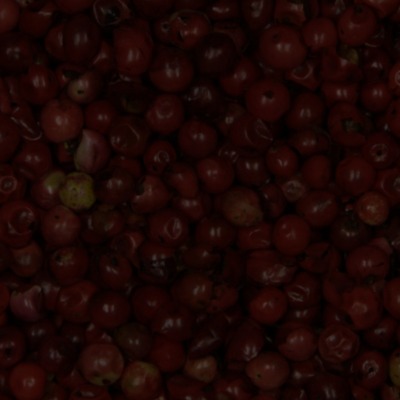} &
        \includegraphics[width=0.1\linewidth]{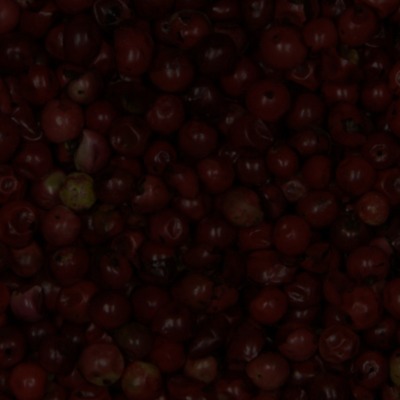} &
    \includegraphics[width=0.1\linewidth]{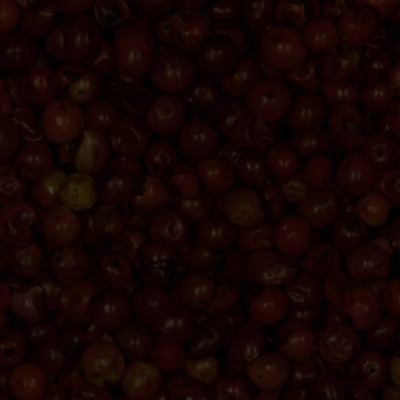} \\[4pt]
        Weighted Gray-Edge &
    \includegraphics[width=0.1\linewidth]{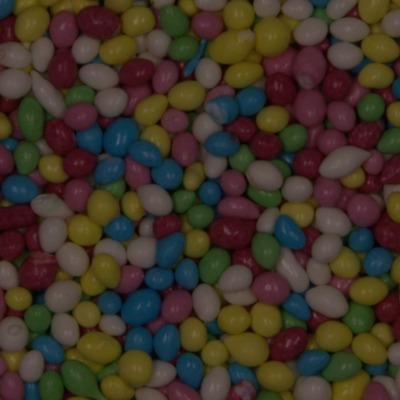} &
        \includegraphics[width=0.1\linewidth]{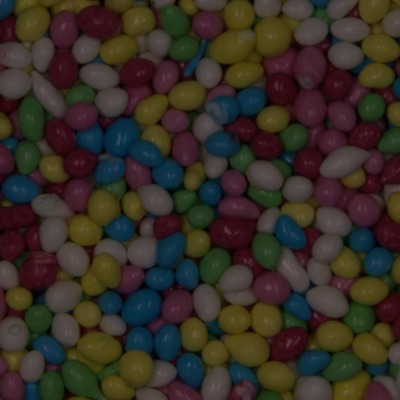} &
    \includegraphics[width=0.1\linewidth]{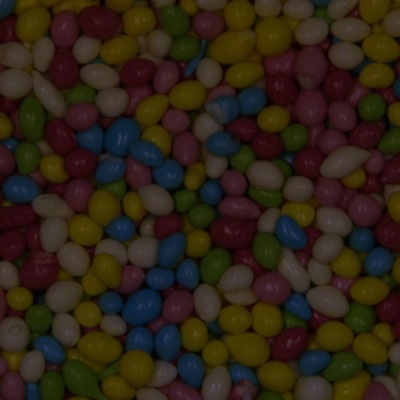} &&
  \includegraphics[width=0.1\linewidth]{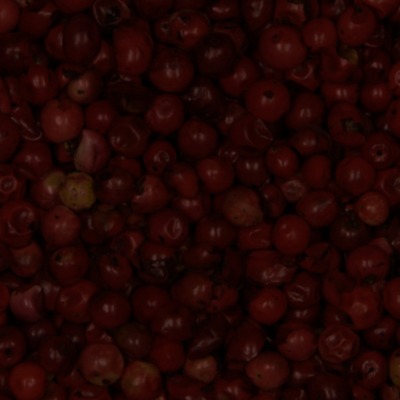} &
        \includegraphics[width=0.1\linewidth]{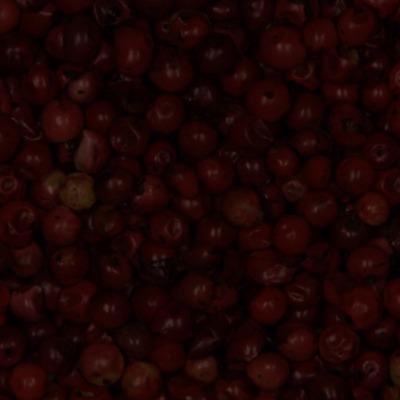} &
    \includegraphics[width=0.1\linewidth]{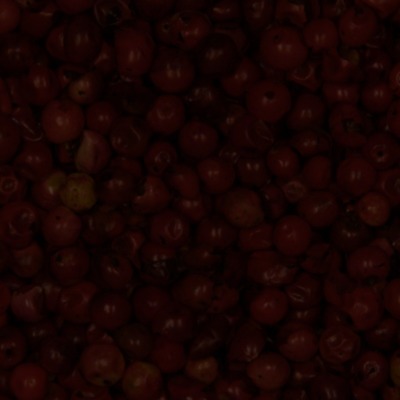} \\
&    D65 ($\theta$=$90^{\circ}$)&D95 ($\theta$=$90^{\circ}$)&L27 ($\theta$=$90^{\circ}$)    &&D65 ($\theta$=$90^{\circ}$)&D95 ($\theta$=$90^{\circ}$)&L27 ($\theta$=$90^{\circ}$)\\
    \end{tabular}
    \caption{Color normalization methods applied to two different
      samples acquired under three different lights.}
  \label{fig:preprocessing}
\end{figure*}

\subsection{Preprocessing with color normalization}

We have preprocessed all the images with five state-of-the-art color
normalization methods. Examples of preprocessing applied to two
different samples are represented in Figure~\ref{fig:preprocessing}.

Table~\ref{tab:normalization_res} reports the performance obtained by
these color normalization methods combined with a selection of
descriptors. It is clear that color normalization helps to improve
performance in the case of CNNs. In particular, the combination of
\emph{Vgg VeryDeep 16} with \emph{Retinex Frankle} achieves an
improvement of 5\% in both the cases of temperature and/or direction
variations. This result confirms the fact that CNNs have been trained
on images without considering changes in illumination conditions. In
contrast, the combination of hand-crafted features with pre-processing
methods, in most of the cases, does not bring any improvements in
terms of classification rate. This is due to the fact that those
features, except for color histogram, have been designed to be more
robust to changes in the temperature of the light.

\section{Summary}
\label{sec:conclusions}

In order to obtain reliable classification of color textures under
uncontrolled conditions, we believe that the descriptors performance
should be assessed under a large set of carefully controlled
variations of lighting conditions.  We described \raw, a database of
texture images acquired under variable light direction, color, and
intensity.  The images of the database will be made publicly available
together with all the scripts used in our experimentation.  We will
also disclose the detailed technical specifications of the hardware
and software used to acquire the database; this will allow the
researchers in this area to extend \raw{} or to acquire their own
database.

\raw{} allowed us to conduct a variety of experiments in which traditional texture, object recognition and CNN-based descriptors have been evaluated in terms
of their capabilities in dealing with single and combined variations
in the lighting conditions.  These experiments made very clear the
strengths and the weaknesses of the investigated approach and clearly
outlined open issues that should be addressed to actually design color
texture descriptors robust with respect to unknown variations in the
imaging conditions.

In extreme summary, we can conclude that:
\begin{itemize}
\item \emph{Traditional texture descriptors} are effective only when
  images have no-variations in lighting conditions.
\item \emph{Object recognition descriptors} demonstrated to perform,
  in most of the cases, better than the \emph{traditional} ones.
\item \emph{CNN-based descriptors} confirmed to be powerful also on
  texture classification tasks outperforming the hand-crafted
  \emph{traditional} and \emph{object-oriented} features.
\item \emph{CNN-based descriptors} handle most of the variations in
  lighting conditions. However for large variations in both color and
  direction of the light, CNN-based descriptors have demonstrated to
  be less effective than object recognition descriptors, especially on
  these classes that are more fine grained.
\item The use of Color normalization did not improve any of the
  hand-crafted descriptors, while for \emph{CNN-based descriptors}
  they demonstrated to be helpful in dealing with complex variations
  in illumination conditions.
\end{itemize}

\bibliographystyle{IEEEtran}
\bibliography{texture}

\end{document}